\documentclass[10pt,twocolumn,letterpaper]{article}

\usepackage[datasets]{wacv}

\usepackage[table]{xcolor}
\usepackage{amsmath,amssymb,amsfonts}
\usepackage{bm}
\usepackage{nicefrac}
\usepackage{microtype}
\usepackage{array}
\usepackage{makecell}
\usepackage{multirow}
\usepackage{tabularx}
\usepackage{booktabs}
\usepackage{comment}
\usepackage{enumitem}
\usepackage{pifont}
\usepackage{wrapfig}
\usepackage{mdframed}

\newcommand{\cmark}{{\fontencoding{TS1}\selectfont\textcolor{black}{\ding{51}}}}
\newcommand{\xmark}{{\fontencoding{TS1}\selectfont\textcolor{black}{\ding{55}}}}

\definecolor{headergray}{gray}{0.9}
\definecolor{sectiongray}{gray}{0.95}
\definecolor{ourrow}{RGB}{230,245,255}

\newcommand{\blu}[1]{\cellcolor{blue!20}#1}
\newcommand{\bblu}[1]{\textbf{\cellcolor{blue!40}#1}}

\newcommand{\second}[1]{\cellcolor{blue!20}#1}
\newcommand{\best}[1]{\textbf{\cellcolor{blue!40}#1}}

\graphicspath{{figure/}{../figure/}}

\definecolor{wacvblue}{rgb}{0.21,0.49,0.74}
\usepackage[pagebackref,breaklinks,colorlinks,allcolors=wacvblue]{hyperref}

\def\wacvPaperID{844} 
\def\confName{WACV}
\def\confYear{2027}

\title{Diagnosing Long-Video Quantitative Reasoning in Multimodal LLMs via Enumeration and Counting}

\author{
Fumihiko Tsuchiya$^{1}$ \quad
Taiki Miyanishi$^{1}$ \quad
Shunsuke Yasuki$^{1}$ \quad
Mahiro Ukai$^{2}$ \quad
Nakamasa Inoue$^{2}$ \\
Shuhei Kurita$^{3}$ \quad
Yusuke Iwasawa$^{1}$ \quad
Yutaka Matsuo$^{1}$ \\[0.5em]
$^{1}$The University of Tokyo, Japan\\
$^{2}$Institute of Science Tokyo, Japan\\
$^{3}$National Institute of Informatics, Japan
}

\begin{document}
\maketitle


\begin{abstract}
Final-answer video QA can show whether a model predicts the right number, but
not which instances it counted, when the supporting evidence occurs, or why it
failed. We diagnose long-video quantitative reasoning in multimodal large
language models (MLLMs) through three coupled abilities: enumerating
query-relevant instances, temporally grounding supporting evidence, and
aggregating the evidence into counts. To support this analysis, we build
EC-Bench, an evidence-annotated evaluation suite with 152 untrimmed videos
longer than 30 minutes, 1{,}699 open-ended queries across six reasoning
categories, and human-verified evidence spans. We evaluate 22 open-source and
proprietary MLLMs using timestamped visual frames and transcripts. The best
average scores reach only 29.98\% Enumeration F1 and 23.74\% Counting accuracy,
compared with human performance of 78.57\% and 82.97\%, respectively. Our
analyses show that counting errors are rarely isolated arithmetic mistakes:
Enumeration F1 is strongly associated with Counting accuracy, temporal grounding
quality is associated with lower counting error, and Counting accuracy drops as
supporting evidence becomes more distributed. These findings recast long-video
counting as evidence retrieval, temporal grounding, deduplication, and
aggregation across the video, rather than simple numerical prediction.
\end{abstract}

\section{Introduction}

\begin{figure*}[!t]
\centering
\includegraphics[width=\linewidth]{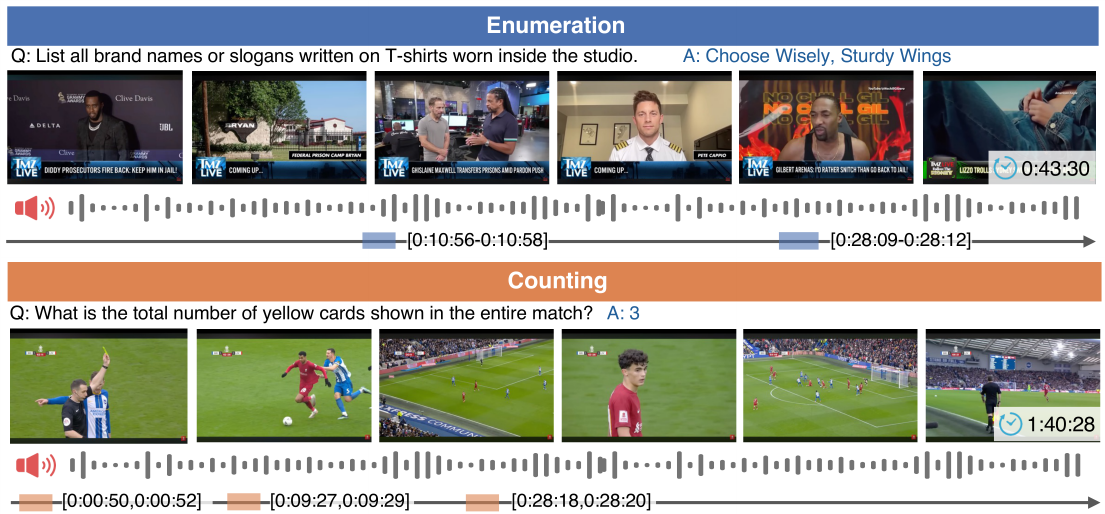}
\caption{%
Example tasks from \textbf{EC-Bench}.
We diagnose long-video quantitative reasoning through three coupled outputs:
\emph{enumeration} of query-relevant instances,
\emph{temporal grounding} of supporting evidence spans,
and \emph{counting} of the identified instances.}
\label{fig:example}
\end{figure*}

Final-answer video question answering (QA) is limited as an evaluation of
quantitative video understanding. A model may predict the correct number without
revealing which instances it counted, when the supporting evidence occurred, or
whether the answer was grounded in the video. This limitation becomes especially
problematic in long-form videos, where relevant events can be sparse, visually
diverse, and separated by tens of minutes. Recent long-video benchmarks have
expanded evaluation to long-form and hour-scale
settings~\cite{hourvideo,longvideobench,lvbench,vrbench}, but final-answer
evaluation alone provides limited visibility into the evidence behind a count.

Counting offers a useful diagnostic setting because the answer is objective, yet
obtaining it requires more than numerical prediction. A model must identify
relevant instances, distinguish them from distractors, avoid duplicate counting
across repeated appearances, and aggregate evidence under the constraints of the
query. Broad video QA and MLLM-oriented benchmarks include counting-related or
temporal reasoning tasks~\cite{tgif-qa,mvbench,seedbench,mmbench}, while
dedicated video counting datasets mainly study repetition, object/event
counting, or audio-visual counting in short, trimmed, or temporally localized
settings~\cite{Countix,dwibedi2024ovrdatasetopenvocabulary,Sinha_2024_ACCV,videoniah,lu2025avreasoner}.
However, these evaluations typically do not diagnose which evidence supports the
model's count.

A central challenge in long-video counting is the gap between local evidence and
global task structure. Individual evidence spans may be brief, but the model
must search an untrimmed video, apply temporal or semantic constraints,
deduplicate repeated or boundary-crossing instances, and produce a consistent
count. Thus, long-video counting tests evidence retrieval, temporal grounding,
deduplication, and aggregation over extended temporal contexts.

Motivated by this view, we formulate long-video quantitative reasoning through
three coupled abilities. \emph{Enumeration} requires listing query-relevant
instances while avoiding irrelevant or duplicate items. \emph{Temporal
grounding} requires localizing the evidence spans that support the answer.
\emph{Counting} requires aggregating the identified evidence into a numerical
answer. This formulation moves beyond final-answer evaluation by making it
possible to inspect what a model counted, where the evidence came from, and
which component failed.

To support this diagnostic study, we construct \textbf{EC-Bench}, an
evidence-annotated evaluation suite for long-video quantitative reasoning.
EC-Bench contains 152 untrimmed videos longer than 30 minutes and 1{,}699
open-ended queries across six reasoning categories, each paired with
human-verified answers and temporal evidence spans. Compared with existing
long-video QA
benchmarks~\cite{egoschema,longvideobench,lvbench,vrbench,hourvideo,chen2024cgbench,Perrett_2025_CVPR}
and video counting
datasets~\cite{Countix,dwibedi2024ovrdatasetopenvocabulary,Sinha_2024_ACCV,videoniah,lu2025avreasoner},
EC-Bench emphasizes open-ended quantitative reasoning with explicit enumeration
and evidence grounding as illustrated in Figure~\ref{fig:example}. This design
complements broad video benchmarks by enabling fine-grained analysis of whether
models identify, localize, deduplicate, and aggregate supporting evidence.

We evaluate 22 open-source and proprietary MLLMs on EC-Bench using timestamped
visual frames and transcripts. The best average scores reach only 29.98\%
Enumeration F1 and 23.74\% Counting accuracy, compared with human performance of
78.57\% and 82.97\%, respectively. Our analyses show that counting failures are
rarely isolated arithmetic mistakes: Enumeration F1 is strongly associated with
Counting accuracy, temporal grounding quality is associated with lower Counting
error, and Counting accuracy drops as supporting evidence becomes more
distributed across the video.

Our contributions are threefold:
\begin{itemize}[leftmargin=10pt]
\item We formulate long-video counting as a coupled problem of instance
enumeration, temporal grounding, and numerical aggregation, moving beyond
final-answer evaluation.
\item We construct EC-Bench, an evidence-annotated diagnostic suite of 152
untrimmed 30+ minute videos and 1{,}699 open-ended queries for evaluating
long-video quantitative reasoning.
\item We conduct a broad diagnostic evaluation of 22 open-source and
proprietary MLLMs, showing that current failures are linked to incomplete
evidence identification, weak temporal grounding, and inconsistent
deduplication across distributed evidence.
\end{itemize}

\setlength{\tabcolsep}{4pt}
\setlength{\aboverulesep}{0pt}
\setlength{\belowrulesep}{0pt}
\setlength{\extrarowheight}{0pt}

\begin{table*}[t]
\centering
\footnotesize
\renewcommand{\arraystretch}{0.95}
\setlength{\tabcolsep}{3pt}
\begin{tabular}{lcccccccc}
\toprule
\textbf{Dataset} & \textbf{\#QA} & \textbf{\#Tasks} & 
\textbf{\makecell{Ultra-long \\ ($\geq$ 30min)}} & 
\textbf{\makecell{Multi- \\ hop}} &
\textbf{\makecell{Evidence \\ Span}} & 
\textbf{Audio} & 
\textbf{Enum.} & 
\textbf{Answer Type} \\
\midrule

\rowcolor{sectiongray}
\multicolumn{9}{l}{{\textbf{Long Video Understanding Benchmarks}}} \\
EgoSchema~\cite{egoschema} & 5{,}063 & 1 & \xmark & \xmark & \xmark & \xmark & \xmark & MCQ \\
ActivityNet\text{-}QA~\cite{activitynetqa} & 58{,}000 & 3 & \xmark & \xmark & \xmark & \xmark & \xmark & MCQ \\
MultiHop\text{-}EgoQA~\cite{mutihopegoqa} & 11{,}707 & 6 & \xmark & \cmark & \cmark & \xmark & \xmark & MCQ \\
HourVideo~\cite{hourvideo} & 12{,}976 & 4 & \cmark & \xmark & \xmark & \xmark & \xmark & MCQ \\
LVBench~\cite{lvbench} & 1{,}549 & 6 & \cmark & \xmark & \xmark & \xmark & \xmark & MCQ \\
LongVideoBench~\cite{longvideobench} & 6{,}678 & 17 & \cmark & \xmark & \xmark & \cmark & \xmark & MCQ \\
VRBench~\cite{vrbench} & 9{,}468 & 7 & \cmark & \cmark & \cmark & \xmark & \xmark & MCQ/Open \\
CG\text{-}Bench~\cite{chen2024cgbench} & 12{,}129 & 3 & \cmark & \cmark & \cmark & \xmark & \xmark & MCQ/Open \\
\midrule

\rowcolor{sectiongray}
\multicolumn{9}{l}{{\textbf{Counting Benchmarks}}} \\
RepCount~\cite{Repcount} & 19{,}280 & 1 & \xmark & \xmark & \xmark & \xmark & \xmark & Numeric \\
Countix~\cite{Countix} & --- & 1 & \xmark & \xmark & \xmark & \xmark & \xmark & Numeric \\
OVR (Open\text{-}Vocab Rep.)~\cite{dwibedi2024ovrdatasetopenvocabulary} & --- & 1 & \xmark & \xmark & \xmark & \xmark & \xmark & Numeric \\
CG-AV-Count~\cite{lu2025avreasoner} & 1{,}027 & 3 & \xmark & \xmark & \cmark & \cmark & \xmark & Numeric \\

\textbf{EC-Bench (Ours)} & \textbf{1{,}699} & \textbf{6} & 
\textbf{\cmark} & \textbf{\cmark} & \textbf{\cmark} & \textbf{\cmark} & \textbf{\cmark} & 
\textbf{Numeric/Open} \\

\bottomrule
\end{tabular}

\caption{Comparison of representative datasets for long-video understanding and counting tasks. EC-Bench uniquely supports ultra-long videos, multi-hop reasoning, explicit evidence spans, and open-ended enumeration.}
\label{tab:comparison}
\end{table*}

\section{Related Work}

\noindent\textbf{MLLMs for Long-Video Understanding.}
Recent multimodal large language models (MLLMs) extend LLMs to visual and video
inputs and achieve strong results on image- and video-language
tasks~\cite{gemini-1.5,gpt4,cogagent,Bai2023QwenVL,llava,instructblip}.
Many video-capable systems represent videos as sampled frames or compressed
visual tokens~\cite{llavanextvideo,internvl,cogvlm2,video-llama,llava_video,qwen2vl}.
Long-form videos introduce a different bottleneck: relevant evidence can be
sparse, repeated, and separated by long temporal gaps.
Prior work has therefore explored visual-token compression, memory or
hierarchical summarization, long-context architectures, and explicit temporal
search or reasoning
strategies~\cite{llama-vid,lvchat,moviechat,malmm,oryx,longva,longvila,pan2026timesearchr,feng2026videor1}.
These works aim to improve temporal coverage and reasoning, but common
evaluations often remain final-answer based.
EC-Bench instead uses enumeration, temporal grounding, and counting to inspect
whether models recover the correct evidence set and aggregate it without
omissions or duplicates.

\noindent\textbf{Benchmarks for Video-Based MLLMs.}
Video QA benchmarks have evolved from short
clips~\cite{tgif-qa,Xu_AAAI17_MSVD-QA,tvqa,next-qa} to broader MLLM-oriented
evaluations~\cite{mvbench,egoschema} and long-form or hour-scale
settings~\cite{activitynetqa,videomme,mlvu,longvideobench,movqa,lvbench,vrbench,hourvideo,chen2024cgbench,Perrett_2025_CVPR}.
These benchmarks are valuable for broad model comparison, as summarized in
Table~\ref{tab:comparison}, but most are not designed to diagnose quantitative
reasoning.
Multiple-choice or final-answer formats reveal whether a model selected or
produced a correct answer, but not whether it identified all supporting
instances.
Even when temporal reasoning or clue grounding is evaluated, models are usually
not required to enumerate every query-relevant instance and aggregate the
recovered set into a count.
EC-Bench focuses on this missing diagnostic layer.

\noindent\textbf{Video Counting and Quantitative Reasoning.}
Video counting has been studied in repetition counting, open-vocabulary
repetition counting, object/event counting, and audio-visual
counting~\cite{Countix,dwibedi2024ovrdatasetopenvocabulary,Sinha_2024_ACCV,videoniah,lu2025avreasoner}.
These datasets provide useful tests of local counting, but they typically
involve short or trimmed clips, or settings where the relevant temporal region
is given or limited.
EC-Bench targets a different failure mode: in an untrimmed long video, a model
must recover the complete set of query-relevant instances, localize their
supporting evidence, remove duplicates across repeated appearances or clip
boundaries, and produce a globally consistent count.
EC-Bench complements existing video QA and counting benchmarks by evaluating
whether models can recover, ground, deduplicate, and aggregate sparse evidence
across full untrimmed videos.

\section{EC-Bench: Diagnostic Evaluation Design}
\label{sec:dataset}

\begin{figure*}[t]
\centering
\includegraphics[width=0.99\linewidth]{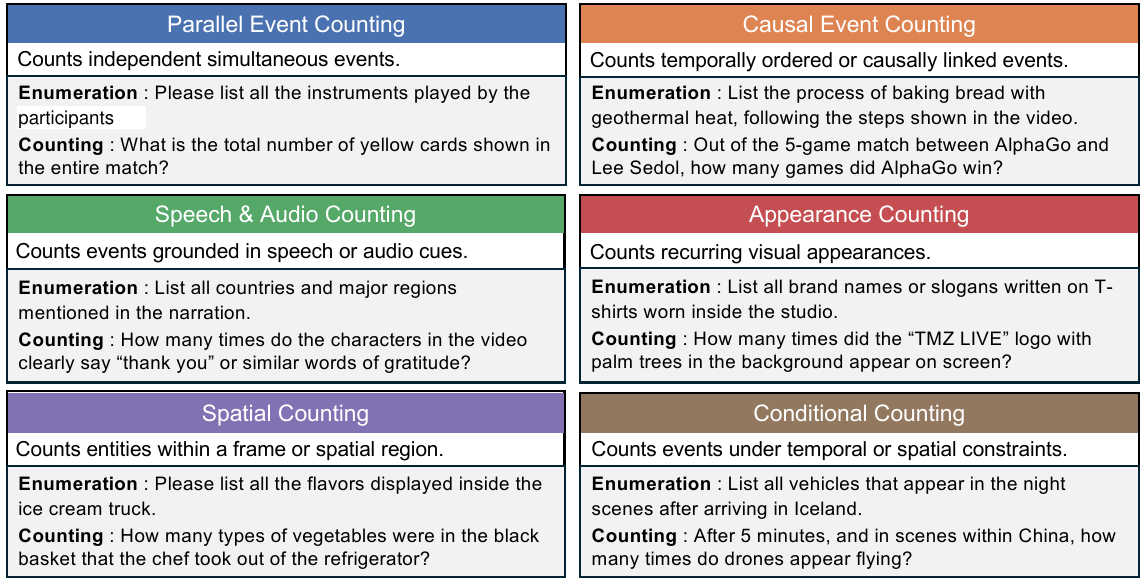}
\vspace{-1em}
\caption{\textbf{Reasoning categories in EC-Bench.}
Examples of six quantitative reasoning types evaluated in the benchmark.}
\vspace{-1em}
\label{fig:reasoning_category}
\end{figure*}

We build \textbf{EC-Bench}, an evidence-annotated diagnostic evaluation suite
for long-video quantitative reasoning.
Rather than evaluating only whether a model outputs the correct final number,
EC-Bench is designed to inspect three coupled abilities: identifying relevant
countable instances, localizing their supporting evidence spans, and aggregating
them into a consistent numerical answer over extended temporal contexts.

For each video-query pair, EC-Bench provides three forms of annotation: an
answer, a set of countable instances when applicable, and one or more temporal
evidence spans on the original video timeline.
A model is evaluated not only by its final answer, but also by whether it
identifies the correct instances and localizes the evidence supporting them.
Individual evidence spans may be short, but they are not provided to the model
in advance; the model must find all relevant evidence across the full untrimmed
video and avoid omissions, duplicates, and unsupported counts.

\subsection{Evaluation Targets}

\noindent\textbf{Enumeration.}
Enumeration requires a model to list all query-relevant instances.
The goal is to evaluate instance-set recovery: a correct output should include
valid instances while avoiding irrelevant or duplicate items.
Enumeration therefore exposes errors that final-answer counting can hide, such
as missing instances, hallucinated items, duplicate items, or incorrect
aggregation across time.

\noindent\textbf{Counting.}
Counting requires a model to aggregate relevant occurrences or instance types
into a single numerical answer under the constraints specified by the query.
Counting is evaluated separately from Enumeration because a model may predict a
plausible number without identifying the correct evidence, or identify some
relevant instances but fail to deduplicate and aggregate them correctly.

\noindent\textbf{Temporal evidence grounding.}
For each query, EC-Bench annotates one or more evidence spans that directly
support the answer.
Models are asked to return evidence intervals along with their answers.
This enables us to evaluate whether the model's output is grounded in the video,
rather than inferred from priors, transcripts alone, or spurious correlations.

\subsection{Reasoning Categories}

EC-Bench covers six reasoning categories that stress different forms of
quantitative video understanding:
Parallel Event Counting, Causal Event Counting, Speech \& Audio Counting,
Appearance Counting, Spatial Counting, and Conditional Counting.
These categories are used to diversify the diagnostic setting rather than to
define disjoint skills.
For example, Appearance and Spatial questions often test local visual
discrimination, whereas Causal and Conditional questions require applying
temporal or semantic constraints across a longer context.
Speech \& Audio questions test whether models can combine transcript or
audio-derived cues with visual evidence.
Figure~\ref{fig:reasoning_category} shows representative examples.

\begin{figure*}[t]
\vspace{-0.5em}
\centering

\begin{subfigure}[t]{0.32\textwidth}
\vspace{0pt}
\centering
\includegraphics[width=\linewidth]{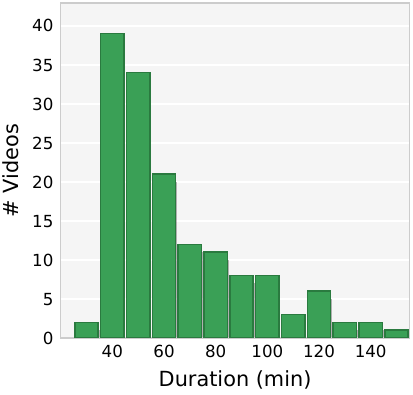}
\caption{Video duration (min).}
\label{fig:stats_video_duration}
\end{subfigure}
\hfill
\begin{subfigure}[t]{0.33\textwidth}
\vspace{0pt}
\centering
\includegraphics[width=\linewidth]{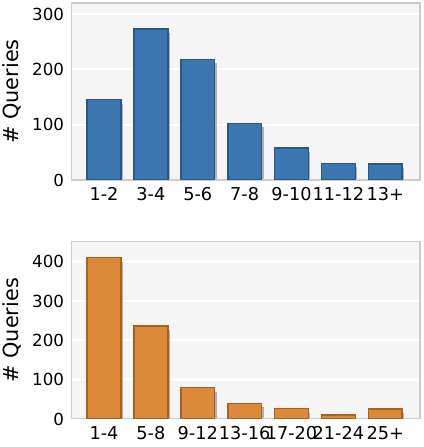}
\caption{Answer distributions: Enumeration (upper) and Counting (lower).}
\label{fig:stats_answer_dist}
\end{subfigure}
\hfill
\begin{subfigure}[t]{0.33\textwidth}
\vspace{0pt}
\centering
\includegraphics[width=\linewidth]{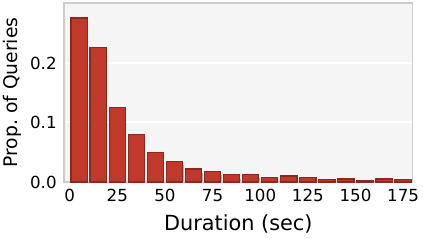}\\[2pt]
\includegraphics[width=\linewidth]{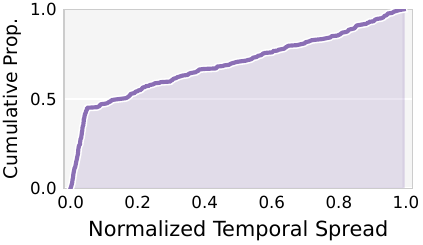}
\caption{Evidence span duration (upper) and normalized temporal spread CDF (lower).}
\label{fig:stats_evidence}
\end{subfigure}

\vspace{-0.5em}

\caption{
\textbf{Dataset statistics.}
\subref{fig:stats_video_duration} Video duration distribution.
\subref{fig:stats_answer_dist} Per-query answer-size distributions for the
Enumeration (upper) and Counting (lower) tasks.
\subref{fig:stats_evidence} Evidence span duration distribution and the
cumulative distribution of normalized temporal spread between the first and
last supporting span.
}
\label{fig:dataset_statistics}
\vspace{-1em}
\end{figure*}

\subsection{Data Construction}
\label{sec:dataset_construction}

EC-Bench contains 152 untrimmed videos, each longer than 30 minutes, with a
median duration of 47 minutes and an average duration of approximately 57
minutes.
The videos are built from LVBench~\cite{lvbench} and additional public sources
across diverse domains, including sports, TV programs, documentaries, live
recordings, cartoons, news, academic lectures, and online self-media.
The final dataset contains 1{,}699 open-ended queries, including 858
Enumeration queries and 841 Counting queries.
We exclude multiple-choice formats to reduce answer-option bias and to require
models to generate answers and supporting evidence directly.

\noindent\textbf{Candidate generation.}
We use \texttt{Gemini-2.5-Pro} to generate initial query-answer-evidence
candidates with a structured prompt.
The prompt asks for objective, video-grounded Enumeration and Counting queries
across the six reasoning categories, requires concise answer formats, and
requests evidence spans.
The generated candidates serve only as drafts; all final annotations are human
verified.
The detailed prompt is provided in Appendix~\ref{app:prompt_qa-gen}.

\noindent\textbf{Human verification and quality control.}
We use a multi-stage human annotation process to refine the generated
candidates.
Annotators remove ambiguous or redundant queries, correct missing or
misidentified instances, revise numerical answers, and adjust evidence span
boundaries to align with the supporting visual or audio events.
The annotation was conducted by 14 annotators from a professional annotation
company, and 60.9\% of the automatically generated QA pairs required
correction, including answer revision, missing-instance correction, duplicate
handling, query filtering, and evidence-boundary adjustment.
Each sample was cross-checked by a second annotator before inclusion in the
benchmark; samples with unresolved ambiguity were removed from the final set.
The annotation guidelines specify answer rules, evidence-boundary rules,
duplicate handling, second-level timestamp precision, and a 5-second merge rule
for repeated appearances.
The full annotation protocol is provided in Appendix~\ref{app:anno-protocol}.

\subsection{Dataset Statistics}
\label{sec:dataset_statistics}

EC-Bench is designed to support diagnostic analysis while maintaining domain
diversity.
The dataset covers a diverse set of video domains, including sports, TV
programs, documentaries, live recordings, cartoons, news, academic lectures,
and online self-media.
This diversity matters because long-video quantitative reasoning can depend on
visual style, event density, audio usage, and narrative structure.
We report the full domain distribution in Appendix~\ref{app:dataset_statistics}.

Figure~\ref{fig:dataset_statistics} summarizes the duration, answer, and
evidence distributions of EC-Bench.
The video duration distribution is right-skewed: while many videos are between
30 and 60 minutes long, the dataset also includes substantially longer videos
exceeding 100 minutes.
The six reasoning categories are approximately balanced across Enumeration and
Counting, with each category accounting for about 16--17\% of the queries.

For Enumeration, the number of target instances is concentrated between 3 and
6, with a mean of 5.32 and a long tail of cases exceeding ten items.
For Counting, most numerical answers fall between 1 and 8, while EC-Bench also
includes larger-count cases that require longer-range aggregation and duplicate
suppression.

The evidence annotations further characterize the long-video challenge.
Each query is supported by 3.48 evidence spans on average, indicating that many
questions require recovering multiple pieces of evidence rather than a single
isolated moment.
At the same time, the relevant evidence is usually brief: the median evidence
duration is about 17 seconds.
However, these short evidence spans are often dispersed over much longer
temporal ranges.
The average temporal spread between the first and last supporting span is
13.72 minutes.
The cumulative distribution of normalized temporal spread in
Figure~\ref{fig:dataset_statistics} shows that, although some queries are
temporally localized, a substantial subset requires retrieving evidence from
distant parts of the video.

\subsection{Evaluation Metrics}
\label{sec:evaluation}

\enlargethispage{1\baselineskip}

\noindent\textbf{Enumeration.}
For Enumeration, we evaluate predicted item lists using precision, recall, and
F1.
Because valid items may be expressed with synonyms or paraphrases, we use an
LLM-as-a-Judge procedure (see Appendix~\ref{app:prompt_llm-judge}) to determine
semantic matches between predicted and ground-truth items.
Matched pairs are counted as true positives (TPs), while unmatched predicted
and ground-truth items are treated as false positives (FPs) and false
negatives (FNs), respectively.
We then compute precision, recall, and F1 in the standard way, and use F1 as
the primary Enumeration metric.
To validate the LLM-as-a-Judge procedure, we conduct a human study on 100
Enumeration queries and observe strong agreement with human-scored F1
(Spearman's $\rho = 0.823$, $p < 0.0001$).

\noindent\textbf{Counting.}
For Counting, we evaluate the final numerical answer using exact-match accuracy
and Mean Absolute Error (MAE). Exact-match accuracy measures whether the
predicted count exactly matches the ground-truth count, while MAE captures the
magnitude of counting errors when predictions are close but not exact.
Model outputs are normalized before evaluation, including punctuation
normalization and extraction of the final numerical answer.

\noindent\textbf{Evidence spans.}
We evaluate temporal evidence grounding using temporal IoU (tIoU) between
predicted and ground-truth evidence regions on the original video timeline.
When multiple evidence spans are present, we collapse the predicted spans into
a single temporal envelope $P$ and the ground-truth spans into a single
temporal envelope $G$.
We then compute $\mathrm{tIoU} = |P \cap G| / |P \cup G|$.
If a model does not output a valid evidence span, its tIoU is set to zero.
We report mean tIoU over all queries.
This envelope-level tIoU is intended as a coarse grounding diagnostic and does
not separately evaluate individual missed or over-generated evidence spans.


\begin{table*}[t]
    \centering
    \resizebox{\textwidth}{!}{
    \begin{tabular}{l | ccccccc | ccccccc}
    \toprule
    
    &
    
    \multicolumn{7}{>{\columncolor{blue!40}}c|}{\textbf{Enumeration}} &
    \multicolumn{7}{>{\columncolor{orange!70}}c}{\textbf{Counting}} \\
    
    
    \cmidrule{2-8}
    \cmidrule{9-15}
    
    \textbf{Model} &
    \rotatebox{60}{Parallel} &
    \rotatebox{60}{Causal} &
    \rotatebox{60}{Speech} &
    \rotatebox{60}{Appear.} &
    \rotatebox{60}{Spatial} &
    \rotatebox{60}{Cond.} &
    \rotatebox{60}{Avg.} &
    \rotatebox{60}{Parallel} &
    \rotatebox{60}{Causal} &
    \rotatebox{60}{Speech} &
    \rotatebox{60}{Appear.} &
    \rotatebox{60}{Spatial} &
    \rotatebox{60}{Cond.} &
    \rotatebox{60}{Avg.} \\
    
    \midrule
    \rowcolor{sectiongray}
    \multicolumn{15}{l}{\textbf{Open-source Models}} \\
    
    LongLLaVA~\cite{longllava}
    & 12.16 & 21.58 & 12.50 & 13.99 & 7.86 & 5.04 & 12.19
    & 4.17 & 11.59 & 5.00 & 5.67 & 6.38 & 11.68 & 7.37 \\
    
    LongVA DPO 7B~\cite{longva}
    & 14.77 & 10.79 & 15.28 & 24.48 & 14.18 & 7.91 & 14.62
    & 9.03 & 10.87 & 5.00 & 3.55 & 11.35 & 13.87 & 8.92 \\
    
    VideoLLaMA3 2B~\cite{videollama3}
    & 19.05 & 2.88 & 15.28 & 20.14 & 8.51 & 4.32 & 11.83
    & 13.19 & 18.12 & 10.00 & 8.51 & 5.67 & 16.06 & 11.89 \\
    
    VideoLLaMA3 7B~\cite{videollama3}
    & 23.13 & 5.76 & 15.71 & 24.26 & 11.28 & 5.97 & 14.48
    & 14.58 & 18.12 & 7.14 & 14.18 & 11.35 & 11.35 & 12.41 \\
    
    LLaVA-Next-Video 7B~\cite{llavanextvideo}
    & 3.36 & 10.07 & 5.59 & 6.34 & 4.20 & 3.60 & 5.50
    & 7.64 & 11.59 & 7.86 & 8.51 & 8.51 & 14.60 & 9.75 \\
    
    LLaVA-Next-Video 34B~\cite{llavanextvideo}
    & 10.74 & 16.55 & 16.67 & 20.83 & 16.78 & 7.19 & 14.80
    & 9.72 & 16.67 & 10.71 & 8.51 & 8.51 & 21.90 & 12.60 \\
    
    InternVideo2.5 8B~\cite{internvideo2.5}
    & 7.38 & 9.35 & 6.29 & 12.68 & 6.99 & 6.52 & 8.20
    & 12.50 & 19.57 & 10.00 & 10.71 & 14.89 & 16.79 & 14.05 \\
    
    mPLUG-Owl3 7B~\cite{mplugowl}
    & 10.07 & 10.07 & 10.49 & 9.72 & 6.99 & 6.47 & 8.98
    & 7.64 & 13.77 & 9.29 & 14.18 & 17.73 & 13.87 & 12.72 \\
    
    VideoChat-Flash 7B~\cite{videochat}
    & 14.63 & 7.89 & 6.84 & 11.76 & 7.56 & 2.65 & 8.65
    & 13.56 & 14.04 & 5.26 & 12.28 & 13.04 & 18.92 & 12.83 \\
    
    MiMo-VL RL 7B~\cite{mimovl}
    & 19.46 & 15.83 & 14.69 & 17.36 & 19.01 & 11.51 & 16.36
    & 13.89 & {21.74} & 7.86 & 10.64 & 14.18 & 9.49 & 12.96 \\
    
    LongVILA-R1 7B~\cite{longvila}
    & 18.79 & 8.63 & 15.38 & 20.14 & 14.79 & 12.23 & 15.07
    & 14.58 & 18.12 & 12.14 & 9.93 & 12.06 & 14.60 & 13.56 \\
    
    LLaVA-OneVision1.5 8B~\cite{llavaonevision}
    & \best{36.91} & 22.30 & {27.27} & 31.94 & 20.28 & 13.04 & 25.47
    & 13.89 & 24.64 & 7.86 & 8.51 & 9.93 & 19.71 & 14.03 \\
    
    Qwen3-VL 8B~\cite{qwen3vl}
    & 22.15 & 19.42 & 15.97 & 26.57 & 19.58 & 12.95 & 19.49
    & 11.11 & 15.22 & 3.57 & 9.22 & 16.31 & 13.14 & 11.41 \\
    
    Qwen3-VL 32B~\cite{qwen3vl}
    & 19.59 & 13.67 & 15.97 & 17.48 & 16.78 & 12.23 & 16.00
    & 11.81 & 19.57 & 8.57 & 12.06 & 18.44 & 19.71 & 14.27 \\
    
    Vamba~\cite{vamba}
    & 10.27 & 10.07 & 10.71 & 15.38 & 6.99 & 4.41 & 9.68
    & 10.42 & 18.84 & 10.71 & 17.73 & 12.06 & 16.06 & 14.27 \\
    
    InternVL3.5 38B~\cite{InternVL3.5_2024}
    & 12.75 & 9.35 & 8.33 & 20.83 & 16.20 & 7.91 & 12.60
    & 9.72 & 23.19 & 9.29 & 15.60 & 9.93 & 22.63 & 14.98 \\
    
    \midrule
    \rowcolor{sectiongray}
    \multicolumn{15}{l}{\textbf{Proprietary Models}} \\
    
    GPT-4o~\cite{gpt4o}
    & \second{33.64} & 12.38 & \second{32.08} & \best{34.91} & 21.70 & {20.19} & 25.90
    & 17.12 & 21.90 & 14.42 & 18.27 & 26.42 & 21.78 & 19.97 \\
    
    GPT-4.1~\cite{gpt41}
    & 17.27 & 15.24 & 23.81 & 24.53 & 17.92 & {20.19} & 19.81
    & 18.02 & \second{33.33} & \best{16.35} & \second{19.23} & \second{27.36} & \second{24.75} & \second{23.14} \\
    
    GPT-5~\cite{gpt5}
    & 22.73 & \best{23.81} & \best{37.74} & \second{33.96} & \best{35.85} & \second{25.96} & \best{29.98}
    & 18.02 & \best{34.62} & 15.38 & \best{20.19} & 26.42 & 21.78 & 22.70 \\
    
    Gemini 2.0 Flash~\cite{gemini20}
    & 25.23 & 21.57 & 27.62 & 23.30 & 25.49 & 19.61 & 23.83
    & 15.60 & 27.45 & 14.00 & 16.67 & 24.27 & 23.23 & 20.16 \\
    
    Gemini 2.5 Flash~\cite{comanici2025gemini25}
    & 26.76 & \second{23.13} & 23.13 & 33.83 & \second{27.21} & \best{27.13} & \second{26.86}
    & \best{21.68} & 27.48 & \second{16.30} & 17.91 & 19.26 & 20.77 & 20.54 \\
    
    Gemini 2.5 Pro~\cite{comanici2025gemini25}
    & 21.10 & 20.95 & {28.16} & 18.45 & 26.92 & {23.53} & 23.16
    & \second{18.35} & \second{33.33} & 14.14 & 17.65 & \best{31.73} & \best{27.27} & \best{23.74} \\
    
    \bottomrule
    \end{tabular}
    }
    \caption{
    Performance of open-source and proprietary MLLMs on EC-Bench.
    Enumeration F1 and Counting accuracy (\%) are reported for six reasoning categories and their overall averages.
    Best and second-best results in each column are highlighted with {\setlength{\fboxsep}{1pt}\colorbox{blue!40}{darker}} and {\setlength{\fboxsep}{1pt}\colorbox{blue!20}{lighter}} shading, respectively.
    }
    \label{tab:enum_count_all}
    \vspace{-1em}
    \end{table*}

\section{Experiments and Diagnostic Analysis}
\label{sec:experiments}

\begin{figure*}[t]
\centering

\begin{minipage}[t]{0.32\linewidth}
\vspace{0pt}
\centering
\includegraphics[width=\linewidth]{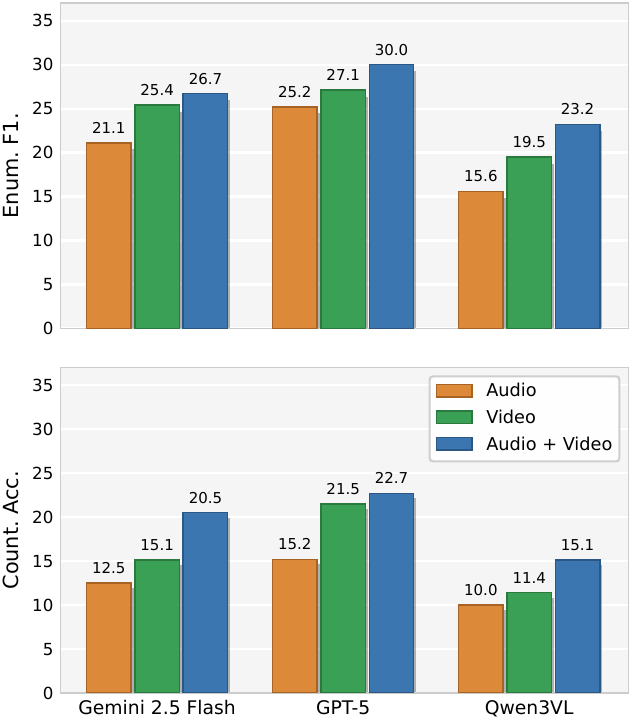}
\captionof{figure}{Modality ablation.}
\label{fig:modality_ablation}
\end{minipage}
\hfill
\begin{minipage}[t]{0.32\linewidth}
\vspace{0pt}
\centering
\includegraphics[width=\linewidth]{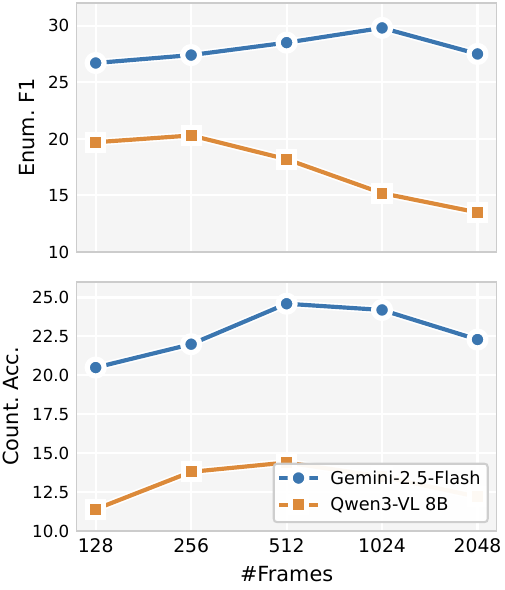}
\captionof{figure}{Impact of frame count.}
\label{fig:frame_ablation}
\end{minipage}
\hfill
\begin{minipage}[t]{0.32\linewidth}
\vspace{0pt}
\centering
{\footnotesize
\setlength{\tabcolsep}{4pt}
\setlength{\aboverulesep}{1pt}
\setlength{\belowrulesep}{1pt}
\renewcommand{\arraystretch}{1.0}
\resizebox{\linewidth}{!}{
\begin{tabular}{lcc}
\toprule
\textbf{Model} & \textbf{MAE}$\downarrow$ & \textbf{Acc.}$\uparrow$ \\
\midrule
\rowcolor{sectiongray}
\multicolumn{3}{l}{\textbf{Open-source Models}} \\
LongLLaVA & 190.29 & 7.37 \\
\hspace{1em} + Enum Prompt & 51.94 & 7.85 \\
LongVA-DPO 7B & 20.46 & 8.92 \\
\hspace{1em} + Enum Prompt & 15.83 & 11.89 \\
\midrule
\rowcolor{sectiongray}
\multicolumn{3}{l}{\textbf{Proprietary Models}} \\
GPT-4o & 5.52 & 19.97 \\
\hspace{1em} + Enum Prompt & 5.16 & 21.20 \\
Gemini-2.5-Flash & 7.63 & 20.54 \\
\hspace{1em} + Enum Prompt & 6.67 & 22.19 \\
\bottomrule
\end{tabular}
}
}
\captionof{table}{Effect of enumeration prompting on counting.}
\label{tab:enum_ablation}
\end{minipage}
\vspace{-1.5em}
\end{figure*}

\subsection{Evaluation Protocol}

\noindent\textbf{Models.}
We evaluate 22 MLLMs: 16 open-source models and six proprietary models, as
listed in Table~\ref{tab:enum_count_all}.
The open-source set spans recent video-capable, long-context, and
reasoning-tuned MLLMs, while the proprietary set includes GPT and Gemini
models.
Together, this model set covers a range of current open-source and proprietary
multimodal systems.

\noindent\textbf{Input protocol.}
To compare models with different input limits, we use a unified
frame-and-transcript protocol with uniformly sampled timestamped frames.
Open-source models use frame counts close to their practical input limits,
while proprietary models use a comparable 128-frame budget; per-model values
are listed in Appendix~\ref{app:input_frame_budget}.
Each frame is paired with its timestamp, allowing models to output evidence
intervals on the original video timeline.
We also provide timestamp-aligned speech transcripts from
Whisper-large-v3~\cite{whisper}.
This is a controlled diagnostic protocol rather than an upper-bound evaluation
for models with native long-video input; we analyze larger frame budgets in
Sec.~\ref{sec:ablation}.

\noindent\textbf{Output protocol.}
All questions are open-ended and contain no multiple-choice options.
For Enumeration, models output a list of query-relevant instances together with
supporting evidence intervals.
For Counting, models output a numerical answer together with supporting
evidence intervals.
Outputs are normalized for punctuation, letter case, and timestamp format
before evaluation.
The detailed prompt and output parsing rules are provided in
Appendix~\ref{app:prompt_qa}.

\subsection{Main Results: Current MLLMs Remain Far Below Humans}
\label{sec:experiments:task}

Table~\ref{tab:enum_count_all} summarizes the quantitative performance of
open-source and proprietary MLLMs on EC-Bench. We report Enumeration F1 and
Counting exact-match accuracy across six reasoning categories, together with
the overall average for each task. To provide a human reference, we conduct a
human evaluation with balanced coverage across the six categories (see
Appendix~\ref{app:human-eval} for details). Human performance reaches 78.57\%
on Enumeration and 82.97\% on Counting.

Current MLLMs remain far below this human reference. The best average
Enumeration score is 29.98\%, achieved by GPT-5, while the best average
Counting score is 23.74\%, achieved by Gemini 2.5 Pro. Thus, even the strongest
models lag human performance by 48.59 percentage points on Enumeration and
59.23 percentage points on Counting.

Proprietary models generally form the strongest group, but the gap differs by
task. For Enumeration, the strongest open-source model, LLaVA-OneVision1.5 8B,
achieves 25.47\%, close to several proprietary models, and obtains the best
score in the Parallel category. For Counting, however, the gap is larger: the
best open-source average is 14.98\%, compared with 23.74\% for the best
proprietary model.

The category-level results show that EC-Bench is difficult across reasoning
types rather than being dominated by a single failure mode. Enumeration scores
are relatively higher in some perception-oriented categories such as Parallel
and Appearance, but remain low overall. Counting exact-match accuracy is also
low across categories; even the best category-level scores for proprietary
models stay in the low-to-mid 30\% range, and Speech counting remains
particularly challenging. These results suggest that long-form quantitative
reasoning requires more than local recognition: models must identify all
relevant instances, localize sparse evidence, avoid duplicate counting, and
aggregate evidence consistently over time.

\subsection{What Drives Long-Video Counting Failures?}

We perform additional analyses to examine the roles of temporal coverage,
visual domain differences, and model capacity.

\noindent\textbf{Counting depends on enumeration.}
Across the 22 models in Table~\ref{tab:enum_count_all}, Enumeration and
Counting show a clear positive association (Spearman's
$\rho = 0.692$, $p < 0.001$): models with higher Enumeration accuracy
tend to achieve higher Counting accuracy.
This suggests that accurate Counting depends not only on numerical prediction,
but also on consistently identifying relevant instances and maintaining
temporal consistency across long videos.

\begin{figure*}[!t]
\centering
\begin{minipage}[t]{0.49\linewidth}
\vspace{0pt}
\centering
{\footnotesize
\setlength{\tabcolsep}{4pt}
\setlength{\aboverulesep}{1pt}
\setlength{\belowrulesep}{1pt}
\renewcommand{\arraystretch}{0.95}
\begin{tabular}{lcccc}
\toprule
\textbf{Model} &
\textbf{\makecell{Prec.$\uparrow$\\(Enum)}} &
\textbf{\makecell{Recall$\uparrow$\\(Enum)}} &
\textbf{\makecell{MAE$\downarrow$\\(Cnt.)}} &
\textbf{tIoU}$\uparrow$ \\
\midrule

\rowcolor{sectiongray}
\multicolumn{5}{l}{\textbf{Open-source Models}} \\

LongLLaVA~\cite{longllava} & 0.22 & 0.10 & 190.29 & 0.00 \\
LongVA DPO 7B~\cite{longva} & 0.26 & 0.13 & 20.46 & 0.01 \\
VideoLLaMA3 2B~\cite{videollama3} & 0.20 & 0.09 & 558.73 & 0.01 \\
VideoLLaMA3 7B~\cite{videollama3} & 0.31 & 0.17 & 43.21 & 0.01 \\
LLaVA-Next-Video 7B~\cite{llavanextvideo} & 0.17 & 0.09 & 26.50 & 0.00 \\
LLaVa-Next-Video 34B~\cite{llavanextvideo} & 0.24 & 0.10 & 10.57 & 0.00 \\
InternVideo2.5 8B~\cite{internvideo2.5} & 0.21 & 0.24 & 5.63 & 0.01 \\
mPLUG-Owl3 7B~\cite{mplugowl} & 0.21 & 0.21 & 32.8 & 0.01 \\
VideoChat Flash 7B~\cite{videochat} & 0.21 & 0.22 & 19.73 & 0.00 \\
MiMo-VL RL 7B~\cite{mimovl} & 0.34 & 0.23 & 40.19 & 0.01 \\
LongVILA R1 7B~\cite{longvila} & 0.31 & 0.17 & 8.88 & 0.00 \\
LLaVA Onevision1.5 8B~\cite{llavaonevision} & 0.35 & 0.13 & 46.14 & 0.01 \\
Qwen3-VL 8B~\cite{qwen3vl} & 0.41 & 0.25 & 29.52 & 0.02 \\
Qwen3-VL 32B~\cite{qwen3vl} & 0.43 & 0.31 & 50.87 & 0.05 \\
Vamba~\cite{vamba} & 0.21 & 0.17 & 550.22 & 0.00 \\
InternVL3.5 38B~\cite{InternVL3.5_2024} & 0.34 & 0.23 & 17.95 & 0.03 \\

\midrule

\rowcolor{sectiongray}
\multicolumn{5}{l}{\textbf{Proprietary Models}} \\

GPT-4o~\cite{gpt4o} & \blu{0.55} & 0.40 & 5.52 & 0.09 \\
GPT-4.1~\cite{gpt41} & 0.47 & 0.56 & \bblu{5.17} & 0.11 \\
GPT-5~\cite{gpt5} & \bblu{0.60} & \blu{0.58} & \blu{5.46} & \blu{0.14} \\
Gemini 2.0 Flash~\cite{gemini20} & 0.52 & 0.41 & 5.77 & 0.11 \\
Gemini 2.5 Flash~\cite{comanici2025gemini25} & 0.52 & 0.56 & 7.63 & 0.08 \\
Gemini 2.5 Pro~\cite{comanici2025gemini25} & \blu{0.55} & \bblu{0.60} & \bblu{5.17} & \bblu{0.16} \\

\bottomrule
\end{tabular}
}
\captionof{table}{Performance with temporal grounding.}
\label{tab:performance_with_tiou}
\end{minipage}%
\hfill%
\begin{minipage}[t]{0.49\linewidth}
\vspace{0pt}
\centering
\includegraphics[width=0.88\linewidth]{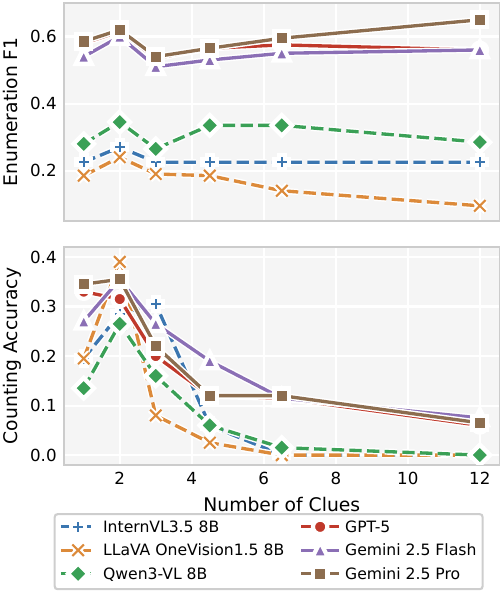}
\captionof{figure}{Performance by clue count.}
\label{fig:perf_by_num_clues}
\end{minipage}
\vspace{-1em}
\end{figure*}

\noindent\textbf{Temporal grounding is associated with quantitative performance.}
Table~\ref{tab:performance_with_tiou} relates temporal evidence localization
(envelope-level tIoU) to quantitative performance.
Models with higher tIoU tend to achieve stronger Enumeration precision and
recall and lower Counting MAE, especially among proprietary models.
For example, GPT-5 and Gemini 2.5 Pro obtain the highest tIoU scores and among
the strongest overall performance.
In contrast, most open-source models show near-zero tIoU, indicating that they
often fail to localize supporting evidence under our evidence-output protocol.
These results suggest that coarse temporal grounding quality is associated with
long-video quantitative performance.

\noindent\textbf{Distributed evidence makes counting difficult.}
Figure~\ref{fig:perf_by_num_clues} plots performance against the number of
evidence clues per query.
Enumeration performance for proprietary models remains relatively stable across
clue counts, while open-source models remain consistently lower.
In contrast, Counting accuracy peaks when only a few clues are required, but
drops sharply as the number of clues increases.
For larger clue counts, several models approach near-zero accuracy, suggesting
that current MLLMs struggle to aggregate distributed evidence even when some
relevant instances are identified.
These results suggest that long-video counting is not only a recognition
problem, but also an evidence aggregation problem over extended temporal
contexts.
A qualitative failure taxonomy in Appendix~\ref{app:failure_taxonomy} further
shows that errors commonly arise from missed evidence, weak grounding,
hallucinated abstractions, and inconsistent deduplication, rather than
isolated arithmetic mistakes.

\subsection{Ablations}\label{sec:ablation}

%

We conduct controlled ablations to examine how input modality, frame density,
and prompting strategy affect long-video quantitative reasoning.

\noindent\textbf{Input modality ablation.}
To verify that EC-Bench requires multimodal reasoning rather than being
solvable from a single modality, we conduct a modality ablation comparing
audio-transcript-only, video-only, and multimodal inputs.
Figure~\ref{fig:modality_ablation} shows that audio-transcript-only inputs
perform worse than video-based settings, indicating that the tasks cannot be
reliably solved from transcripts alone.
Video-only inputs achieve consistently higher accuracy, confirming that visual
evidence is essential for Enumeration and Counting.
Finally, combining video and audio yields the best performance, suggesting
that audio provides complementary cues.
These results indicate that both visual and auditory information contribute to
solving EC-Bench tasks.

\noindent\textbf{Effect of frame density.}
Figure~\ref{fig:frame_ablation} analyzes the impact of input frame count on
model performance.
For Gemini-2.5-Flash, increasing the number of frames improves both
Enumeration and Counting accuracy at moderate frame budgets, indicating that
broader temporal coverage helps reduce missed evidence.
However, the gains diminish at higher frame counts, and performance eventually
decreases, suggesting that simply increasing frame density can introduce
redundancy or noise without reliably improving instance identification.
A similar pattern is observed for Qwen3-VL 8B, where moderate frame increases
provide small improvements but larger frame budgets lead to performance drops.
These results suggest that while additional frames can improve temporal
coverage, naive increases in frame density do not consistently improve
performance.
Instead, performance is likely limited by the model's ability to identify and
integrate relevant instances over long temporal contexts.

\noindent\textbf{Enumeration-first prompting.}
To investigate the impact of Enumeration on Counting, we introduce an explicit
enumeration-first prompting strategy.
In this setup, models are instructed to first enumerate relevant instances
before performing the counting step.
The detailed prompt is provided in Appendix~\ref{app:enum-first-prompt}.
Table~\ref{tab:enum_ablation} evaluates this strategy on two open-source and
two proprietary models.
As shown in Table~\ref{tab:enum_ablation}, enumeration-first prompting
consistently improves Counting accuracy and reduces MAE across all evaluated
models.
The gains are larger for open-source models, while proprietary models show
smaller but consistent improvements.
For LongLLaVA, however, the MAE drop is not matched by accuracy
($7.37 \to 7.85$), suggesting the gain mainly reflects outlier suppression
rather than fundamentally correct counts.
These results suggest that explicitly enumerating instances before aggregation
helps stabilize numerical reasoning in long-form videos and improves Counting
performance without additional model training.

\section{Conclusion}

We presented EC-Bench, an evidence-annotated diagnostic evaluation suite for
long-video quantitative reasoning in multimodal LLMs. Rather than evaluating
only final answers, EC-Bench jointly evaluates three coupled abilities:
Enumeration, temporal evidence grounding, and Counting. Across 22 open-source
and proprietary MLLMs, current systems remain far below humans: the best
average scores reach only 29.98\% Enumeration F1 and 23.74\% Counting accuracy,
compared with human performance of 78.57\% and 82.97\%, respectively.

Our analyses show that counting errors are rarely isolated numerical mistakes:
Enumeration performance, temporal localization quality, and Counting accuracy
are closely related, and performance drops as supporting evidence becomes more
distributed across the video. These findings recast long-video counting as
exhaustive evidence retrieval, temporal grounding, deduplication, and
aggregation over the full video, rather than simple numerical prediction.

{
    \small
    \bibliographystyle{ieeenat_fullname}
    \bibliography{mybib}
}

\clearpage
\appendix

\setlength{\tabcolsep}{4pt}
\setlength{\aboverulesep}{0pt}
\setlength{\belowrulesep}{0pt}
\setlength{\extrarowheight}{0pt}

\begin{figure*}[h]
\centering
\includegraphics[width=0.95\linewidth]{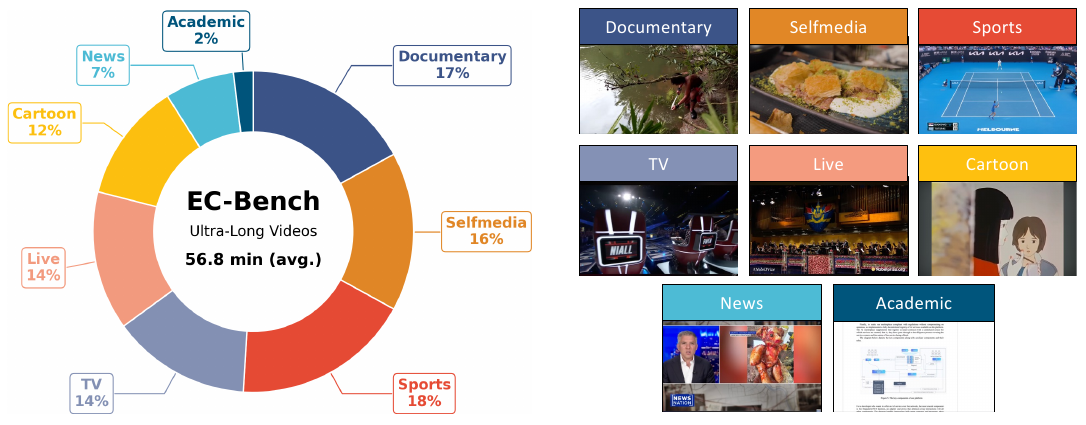}
\caption{\textbf{Video statistics of EC-Bench.}
Left: distribution of video durations (all videos exceed 30 minutes; median
47 minutes).
Right: representative examples of video genres in the dataset.}
\label{fig:video_statistics}
\end{figure*}

\section{Code and Dataset Availability}
The anonymous code repository used for reproducibility during review is
available at:
\url{https://anonymous.4open.science/r/EC-Bench-7DF3}

\section{Limitations}

EC-Bench is intended as a diagnostic evaluation suite, not a comprehensive
benchmark for all long-video understanding abilities. Its scale reflects the
cost of constructing open-ended queries with human-verified answers and
temporal evidence annotations. Although EC-Bench covers diverse video domains
and six reasoning categories, it does not exhaustively cover all genres,
languages, visual styles, or real-world long-video applications.

Our main comparison uses a controlled frame-and-transcript protocol to support
evaluation across heterogeneous models with different input constraints. This
protocol enables fair comparison, but it should not be interpreted as an
upper-bound evaluation for systems with native long-video input or specialized
video retrieval mechanisms. Because videos are represented by uniformly
sampled frames, short or sparsely occurring evidence may be missed by the
input itself, especially when relevant evidence spans are brief and
distributed across long videos.

Our evaluation also has metric-level limitations. The temporal grounding
metric measures overlap between predicted and ground-truth evidence regions,
but it does not fully diagnose every individual missed or over-generated
evidence span. The envelope-level reduction also leaves room for a model
to inflate its tIoU by emitting a single wide interval that brackets all
ground-truth spans together with their inter-span gaps; we therefore plan
to complement envelope tIoU with span-level precision/recall in future
work. In addition, Enumeration is evaluated using an LLM-as-a-Judge
procedure validated against human judgments on 100 queries; while the
validation shows strong agreement, LLM-based semantic matching may still be
sensitive to judge prompts, model choice, and ambiguous paraphrases. Finally,
the human reference is intended as a performance anchor rather than a complete
study of human variability.

\noindent\textbf{Data distribution, licensing, and intended use.}
EC-Bench is distributed as URL-anchored annotations rather than as
video files; users obtain the underlying videos from their original
sources (LVBench~\cite{lvbench} and additional public hosts, primarily
YouTube) under each source's own terms. The annotation data
(queries, evidence intervals, answers) is released under
CC-BY-SA-4.0. We do not redistribute video frames, audio, or
transcripts. EC-Bench is intended as a research benchmark for
evaluating multimodal LLMs on long-video quantitative reasoning;
it is not intended for surveillance, person identification, or
face- or voice-recognition training. Because all videos are
sourced from publicly available material that does not contain
newly captured personal data, we follow the same research-use
position adopted by prior long-video benchmarks (e.g., Kinetics,
LVBench). A datasheet following~\cite{gebru2021datasheets} will be
released with the camera-ready version.

\section{Benchmark Details}\label{sec:benchmark_details}

\subsection{Video Statistics}\label{app:dataset_statistics}
Figure~\ref{fig:video_statistics} presents the distribution of video durations
and genres.
The dataset spans diverse domains including documentaries (17\%), Selfmedia
(16\%), Sports (18\%), TV and Live (14\% each), Cartoons (12\%), News (7\%),
and Academic (2\%).
This diversity in both tasks and video sources allows EC-Bench to evaluate
long-range temporal reasoning and quantitative understanding across varied
video contexts.

\subsection{Human Performance Evaluation}
\label{app:human-eval}

To provide a reference point for model performance on EC-Bench, we conducted a
human evaluation on a subset of benchmark queries. We randomly sampled
117
queries from the dataset with balanced coverage across the six reasoning
categories and both task types. Human evaluators answered each query by
watching the corresponding video and following the same answer format used in
the benchmark.

For Counting, we used exact-match accuracy: an answer was considered correct
only when the predicted number exactly matched the ground-truth value. For
Enumeration, we evaluated human-produced item lists using the same semantic
matching protocol as model outputs. Specifically, predicted and ground-truth
items were matched to compute precision, recall, and F1, and we report
macro-averaged Enumeration F1 over the evaluated queries.

Under this protocol, human performance reached 78.57\% Enumeration F1 and
82.97\% Counting exact-match accuracy. These results provide a useful
reference point for interpreting model performance on EC-Bench. The relatively
lower human scores reflect the inherent difficulty of long-video reasoning,
which often requires identifying multiple instances across extended temporal
contexts.

\begin{table*}[t]
\centering
\resizebox{\textwidth}{!}{
\begin{tabular}{l | cccccccc c| cccccccc c}

\toprule


& \multicolumn{9}{>{\columncolor{blue!40}}c|}{\textbf{Enumeration}} &
\multicolumn{9}{>{\columncolor{orange!70}}c}{\textbf{Counting}} \\

\cmidrule{2-9}
\cmidrule{10-19}

\textbf{Model} &
\rotatebox{70}{ Academic} &
\rotatebox{70}{ Cartoon} &
\rotatebox{70}{ Documentary} &
\rotatebox{70}{ Live} &
\rotatebox{70}{ News} &
\rotatebox{70}{ Self-media} &
\rotatebox{70}{ Sports} &
\rotatebox{70}{ TV} &
\rotatebox{70}{ Average} &
\rotatebox{70}{ Academic} &
\rotatebox{70}{ Cartoon} &
\rotatebox{70}{ Documentary} &
\rotatebox{70}{ Live} &
\rotatebox{70}{ News} &
\rotatebox{70}{ Self-media} &
\rotatebox{70}{ Sports} &
\rotatebox{70}{ TV} &
\rotatebox{70}{ Average} \\

\midrule
\rowcolor{sectiongray}
\multicolumn{19}{l}{\textbf{Open-source Models}} \\

LongLLaVA~\cite{longllava} 
& 10.00 & 16.35 & 12.24 & 16.67 & 13.56 & 8.21 & 13.79 & 6.45 & 11.96
& 9.09 & 7.62 & 6.25 & 6.14 & 8.47 & 9.09 & 4.70 & 10.34 & 7.37 \\

LongVA DPO 7B~\cite{longva}
& 5.00 & \second{18.27} & 17.01 & 9.17 & 20.34 & 12.78 & 16.38 & 7.53 & 14.62
& 0.00 & 9.52 & 4.86 & 11.40 & 13.56 & 6.82 & 8.72 & 12.93 & 8.92 \\

VideoLLaMA3 2B~\cite{videollama3}
& 25.00 & 16.99 & 20.75 & 14.62 & 20.69 & 13.11 & 16.38 & 11.83 & 12.06
& 9.09 & 9.52 & 9.72 & 13.16 & 15.25 & 11.36 & 11.41 & 15.52 & 11.89 \\

VideoLLaMA3 7B~\cite{videollama3}
& 20.00 & 13.26 & 19.05 & 13.89 & 20.69 & 13.11 & 14.66 & 12.90 & 14.96
& 13.64 & 11.43 & 10.42 & 13.16 & 23.73 & 9.85 & 12.08 & 16.38 & 12.96 \\

LLaVA-Next-Video 7B~\cite{llavanextvideo}
& 0.00 & 9.62 & 9.52 & 9.72 & 8.05 & 8.28 & 15.52 & 7.53 & 5.96
& 13.64 & 14.29 & 5.56 & 13.16 & 5.08 & 12.12 & 8.05 & 8.62 & 9.75 \\

LLaVA-Next-Video 34B~\cite{llavanextvideo}
& 10.00 & 16.35 & 20.41 & 16.67 & 20.34 & 15.64 & 21.55 & 14.52 & 14.69
& 4.55 & \second{19.05} & 9.03 & 13.16 & 13.56 & 12.12 & 12.75 & 12.07 & 12.60 \\

InternVideo2.5 8B~\cite{internvideo2.5}
& 11.90 & 12.56 & 10.65 & 11.54 & 13.56 & 6.77 & 13.18 & 10.83 & 8.20
& 18.18 & 13.33 & 11.81 & 14.04 & 16.95 & 9.85 & 16.22 & 17.24 & 14.05 \\

mPLUG-Owl3 7B~\cite{mplugowl}
& 10.00 & 13.46 & 14.29 & 13.89 & 12.71 & 11.15 & 18.97 & 12.90 & 9.33
& 18.18 & 10.48 & 10.42 & 7.02 & 20.34 & 12.12 & 10.07 & 22.41 & 12.72 \\

MiMo-VL RL 7B~\cite{mimovl}
& 21.43 & 11.96 & 14.14 & 14.10 & 21.19 & 11.28 & 17.79 & 13.33 & 16.24
& 18.18 & 15.24 & 13.89 & 11.40 & 15.25 & 12.12 & 10.74 & 12.93 & 12.96 \\

\midrule
\rowcolor{sectiongray}
\multicolumn{19}{l}{\textbf{Proprietary Models}} \\

GPT-4o~\cite{gpt4o}
& {27.78} & \best{18.54} & \second{23.94} & 25.00 & 26.32 & 18.72 & 24.58 & 22.67 & 25.43
& 20.00 & 13.33 & 16.41 & 22.06 & \best{31.25} & 18.35 & 22.50 & 21.92 & 19.97 \\

GPT-4.1~\cite{gpt41}
& 22.22 & 17.88 & 21.24 & \second{27.94} & 21.05 & 19.18 & 23.33 & 19.46 & 19.65
& 20.00 & \best{22.67} & \second{20.31} & \best{27.94} & 27.08 & 20.18 & 24.17 & 24.66 & \second{23.14} \\

GPT-5~\cite{gpt5}
& {27.78} & 15.23 & \best{26.74} & \best{33.09} & \best{34.74} & \second{23.29} & \best{30.42} & \second{24.00} & \best{30.14}
& {30.00} & 10.67 & \best{21.26} & \best{27.94} & 29.17 & \second{22.94} & \second{25.83} & 21.92 & 22.70 \\

Gemini 2.5 Flash~\cite{comanici2025gemini25}
& \best{50.00} & 16.35 & 21.19 & 26.91 & \second{29.57} & 20.08 & 24.25 & \best{25.55} & \second{26.73}
& \second{45.45} & 13.33 & 19.40 & 18.18 & \second{30.51} & 16.67 & 20.00 & \second{26.13} & 20.54 \\

Gemini 2.5 Pro~\cite{comanici2025gemini25}
& \best{50.00} & 17.01 & 17.97 & 27.82 & 26.60 & \best{24.65} & \second{28.51} & 21.68 & 23.48
& \best{50.00} & 17.33 & 15.87 & {24.62} & 29.17 & \best{23.58} & \best{29.31} & \best{27.54} & \best{23.74} \\

\bottomrule
\end{tabular}
}

\caption{
Performance of open-source and proprietary MLLMs on EC-Bench across video genres.
Enumeration F1 and Counting accuracy (\%) are reported for eight video genres and their overall averages.
Best and second-best results in each column are highlighted with {\setlength{\fboxsep}{1pt}\colorbox{blue!40}{darker}} and {\setlength{\fboxsep}{1pt}\colorbox{blue!20}{lighter}} shading, respectively.
}

\label{tab:genre_enum_count}
\end{table*}

\subsection{Annotation Protocol}
\label{app:anno-protocol}
Figure~\ref{fig:anno-protocol} illustrates the detailed workflow of our
long-video annotation pipeline.
The annotation protocol consists of feasibility checking, answer verification,
and evidence extraction. Annotators verify whether each query is answerable,
revise incorrect or ambiguous answers, and extract supporting evidence
intervals with second-level timestamp precision. Additional rules handle
duplicate appearances, type-based counting, and ambiguous cases to ensure
annotation consistency.

\subsection{Prompt Design for QA Generation}\label{app:prompt_qa-gen}
We specify explicit answer rules to ensure consistent annotations.
Enumeration questions require answers as comma-separated lists of entities or
locations, whereas Counting questions require a single numerical value.
To standardize outputs, the prompt enforces a JSON-only format and provides
template examples for multiple video genres (e.g., documentary, news, sports,
and cartoon).
Figure~\ref{fig:qa_generation_prompt} shows the prompt used to generate
enumeration and counting queries and their corresponding answers.

\subsection{Robustness to the QA-Generation Model}\label{app:qa-generation-bias}

Because EC-Bench uses Gemini 2.5 Pro as the seed QA generator (see
Sec.~\ref{app:prompt_qa-gen}), an important concern is whether the
resulting evaluation set systematically advantages models from the
Gemini family. We argue that the observed results are inconsistent with
such a systematic bias, on the following grounds.

\noindent\textbf{(i) Open-ended task leaders are mixed across families.}
On the main results in Table~\ref{tab:enum_count_all}, GPT-5 obtains the
best overall Enumeration F1 (29.98\%) by a clear margin of $+3.12$
points over the second-best (Gemini 2.5 Flash, 26.86\%), while Gemini
2.5 Pro itself is only sixth in Enumeration (23.16\%). For Counting,
the top three averages (Gemini 2.5 Pro 23.74\%, GPT-4.1 23.14\%, GPT-5
22.70\%) lie within a $1.04$-point band that spans both Gemini and GPT
families. A QA-generator bias that systematically favors the generator
family would be expected to place the generator model at or near the
top on both tasks; the actual rankings do not show this pattern.

\noindent\textbf{(ii) Category-level winners are diverse.}
Across the 12 category-level columns (six categories $\times$ two tasks),
Gemini 2.5 Pro is the best model in only two columns
(Counting/Spatial 31.73\% and Counting/Conditional 27.27\%), whereas
the GPT family leads five Counting categories (e.g., Speech: GPT-4.1
16.35\%; Causal: GPT-5 34.62\%; Appearance: GPT-5 20.19\%), Gemini 2.5
Flash wins one (Counting/Parallel 21.68\%), and an open-source model
takes the top slot on Enumeration/Parallel (LLaVA-OneVision1.5 8B,
36.91\%).

\noindent\textbf{(iii) Human verification removes most surface-level
artifacts.} As described in Sec.~\ref{app:anno-protocol}, $60.9\%$ of
auto-generated QA pairs required human correction, including answer
revision, missing-instance correction, duplicate handling, query
filtering, and evidence-boundary adjustment. Surface artifacts
specific to the Gemini family (e.g., recurring phrasings or salience
preferences) are therefore expected to be substantially diluted in the
final benchmark.

Together, these observations indicate that EC-Bench evaluation is not
dominated by an outlier preference for the QA-generation family.
Fully eliminating such a bias would ideally require regenerating
subsets of QA pairs with alternative seed models, which we leave to
future work.

\subsection{Enumeration as a Distinct Diagnostic Axis}\label{app:enumeration-distinctness}

A natural question is whether Enumeration F1 is a relabeled form of
recall that overlaps with Counting accuracy or with prior evidence-span
scoring (e.g., CG-Bench~\cite{chen2024cgbench},
VRBench~\cite{vrbench}). Two observations indicate that Enumeration is
a complementary diagnostic axis.

First, the rank correlation between Enumeration F1 and Counting
accuracy across the 22-model main comparison is Spearman
$\rho = 0.692$ (Sec.~\ref{sec:experiments:task}), well below the
$\rho \approx 1$ that would be expected if Enumeration were a simple
restatement of Counting. Concretely, GPT-5 ranks 1st on Enumeration
(29.98\%) but 3rd on Counting (22.70\%), while Gemini 2.5 Pro ranks
6th on Enumeration (23.16\%) but 1st on Counting (23.74\%), and
GPT-4.1 ranks 7th on Enumeration (19.81\%) but 2nd on Counting
(23.14\%). A model can therefore enumerate well while counting poorly
(aggregation failure) or count well while enumerating poorly
(numerical guesswork without evidence retrieval).

Second, prior benchmarks evaluate evidence-span retrieval as a binary
test on multiple-choice or single-answer outputs and do not require
the model to emit an open-ended set of relevant instances. As a
result, prior metrics cannot diagnose missing-instance, duplicate, or
hallucinated-instance failures within a single query; these failure
modes are first-class outcomes in our protocol and are illustrated in
the qualitative failure taxonomy (Sec.~\ref{app:failure_taxonomy}).

\subsection{Per-model Input Frame Budget}\label{app:input_frame_budget}

Table~\ref{tab:frame_budget} lists the input frame budget used for each
model in our evaluation. The default budget is 128 uniformly sampled
frames. Three open-source models use a smaller budget set to their
model-specific upper bounds: LLaVA-Next-Video 7B/34B (64 frames) and
MiMo-VL RL 7B (80 frames). The InternVL3.5 8B variant used in the
size-matched spotlight analyses (Sec.~\ref{sec:additional_results}) uses
50 frames due to its shorter context window.

\begin{table}[h]
\centering
\small
\begin{tabular}{lc}
\toprule
\textbf{Model} & \textbf{\#Frames} \\
\midrule
\multicolumn{2}{l}{\textit{Open-source Models}} \\
LongLLaVA & 128 \\
LongVA DPO 7B & 128 \\
VideoLLaMA3 2B & 128 \\
VideoLLaMA3 7B & 128 \\
LLaVA-Next-Video 7B & 64 \\
LLaVA-Next-Video 34B & 64 \\
InternVideo2.5 8B & 128 \\
mPLUG-Owl3 7B & 128 \\
VideoChat-Flash 7B & 128 \\
MiMo-VL RL 7B & 80 \\
LongVILA-R1 7B & 128 \\
LLaVA-OneVision1.5 8B & 128 \\
Qwen3-VL 8B & 128 \\
Qwen3-VL 32B & 128 \\
Vamba & 128 \\
InternVL3.5 38B & 128 \\
\midrule
\multicolumn{2}{l}{\textit{Proprietary Models}} \\
GPT-4o & 128 \\
GPT-4.1 & 128 \\
GPT-5 & 128 \\
Gemini 2.0 Flash & 128 \\
Gemini 2.5 Flash & 128 \\
Gemini 2.5 Pro & 128 \\
\midrule
\multicolumn{2}{l}{\textit{Spotlight variant (Sec.~\ref{sec:additional_results})}} \\
InternVL3.5 8B & 50 \\
\bottomrule
\end{tabular}
\caption{Per-model input frame budget.}
\label{tab:frame_budget}
\end{table}

\begin{table*}[t]
\centering
\footnotesize
\setlength{\tabcolsep}{4pt}
\renewcommand{\arraystretch}{1.15}

\definecolor{tablegray}{RGB}{248,248,248}
\definecolor{tblheadergray}{RGB}{235,235,235}

\rowcolors{2}{tablegray}{white}

\begin{tabular}{p{0.14\linewidth}|p{0.42\linewidth}|p{0.32\linewidth}}
\toprule
\rowcolor{tblheadergray}
\textbf{Failure type} & \textbf{Description} & \textbf{Representative example} \\
\midrule

\textbf{Missed instances}
& The model captures the coarse scene but fails to identify all relevant
people, objects, or events, particularly under sparse or crowded evidence
settings.
& For ``What is the maximum number of people trapped in capsules?'', the
ground truth is 87, while GPT-4o predicts 3. \\

\midrule

\textbf{Weak temporal grounding}
& The answer is semantically plausible, but the predicted evidence spans
fail to localize all supporting events.
& Retrieved intervals are missing, overly broad, or misaligned with the
actual evidence. \\

\midrule

\textbf{Transcript grounding failure}
& The model relies on topic-level transcript or audio cues instead of
counting concrete visual occurrences.
& For ``How many times is AI-generated content shown?'', the model predicts
5 instead of the correct count of 26. \\

\midrule

\textbf{Over-abstraction}
& The model outputs coarse semantic categories instead of enumerating the
required concrete instances.
& For ``List the technical terms for Changi Airport T4'', specific terms
are replaced by broad labels such as ``self-service systems''. \\

\midrule

\textbf{Prior-knowledge hallucination}
& The model injects plausible world knowledge unsupported by video evidence.
& For ``List the items under `GENERATIVE AI' at 6:54'', the model
hallucinates NVIDIA-related terms. \\

\midrule

\textbf{Deduplication failure}
& The model inconsistently merges or splits repeated appearances across
long temporal contexts.
& Boundary-crossing or repeated events are counted inconsistently across
segments. \\

\bottomrule
\end{tabular}

\caption{
\textbf{Failure taxonomy on EC-Bench.}
Representative GPT-4o failures grouped by the stage of long-video
quantitative reasoning where the error occurs.
}
\label{tab:failure_taxonomy}
\end{table*}

\subsection{Prompt Design for Long-Video QA}\label{app:prompt_qa}
As shown in Figure~\ref{fig:qa_prompt}, we use a structured multimodal
prompting framework designed specifically for long-video enumeration and
counting.
The prompt defines strict rules for numerical answers, itemized enumerations,
and temporal evidence extraction, and instructs models to jointly use visual
frames and audio data.
This design enforces a consistent answering protocol across models when
generating responses and selecting supporting evidence clips.
Figure~\ref{fig:enum_first_prompt} also shows the Enumeration-First Prompting
strategy.

\subsection{Prompt Design for LLM-based Evaluation}\label{app:prompt_llm-judge}
Figure~\ref{fig:llm_judge_prompt} shows the prompt used for the LLM-as-Judge
evaluation protocol for the Enumeration task.
The prompt instructs the model to first produce the most accurate short answer
based solely on the video content, and then identify all supporting evidence
spans.
It specifies strict rules for timestamp formats, evidence localization, and
output structure, enabling consistent and reproducible evaluation across
models.

\subsection{Enumeration-First Prompting}\label{app:enum-first-prompt}
Figure~\ref{fig:enum_first_prompt} shows the enumeration-first prompting
template used in our ablation study.
The prompt instructs the model to first enumerate all query-relevant
instances, describe each instance with distinguishing details, deduplicate
repeated or overlapping instances, and then compute the final numerical
answer.
It also requires the model to output supporting evidence clips for each
answer, enabling us to test whether explicit instance enumeration improves
counting performance without additional training.

\begin{figure*}[t]
  \centering
  \begin{subfigure}[c]{0.48\linewidth}
    \centering
    \includegraphics[width=\linewidth]{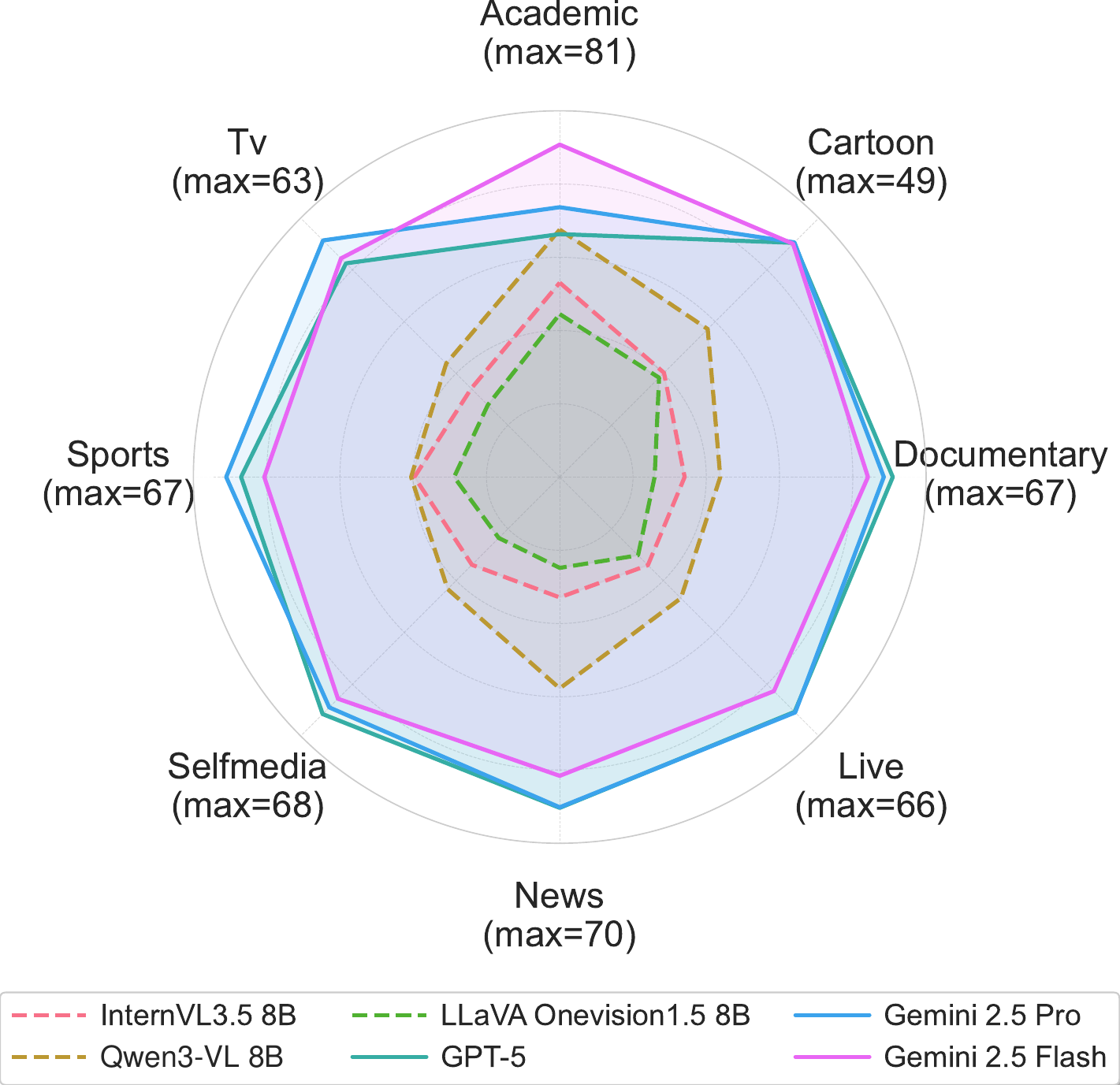}
    \caption{Enumeration.}
    \label{fig:radar_enum_chart}
  \end{subfigure}
  \begin{subfigure}[c]{0.48\linewidth}
    \centering
    \includegraphics[width=\linewidth]{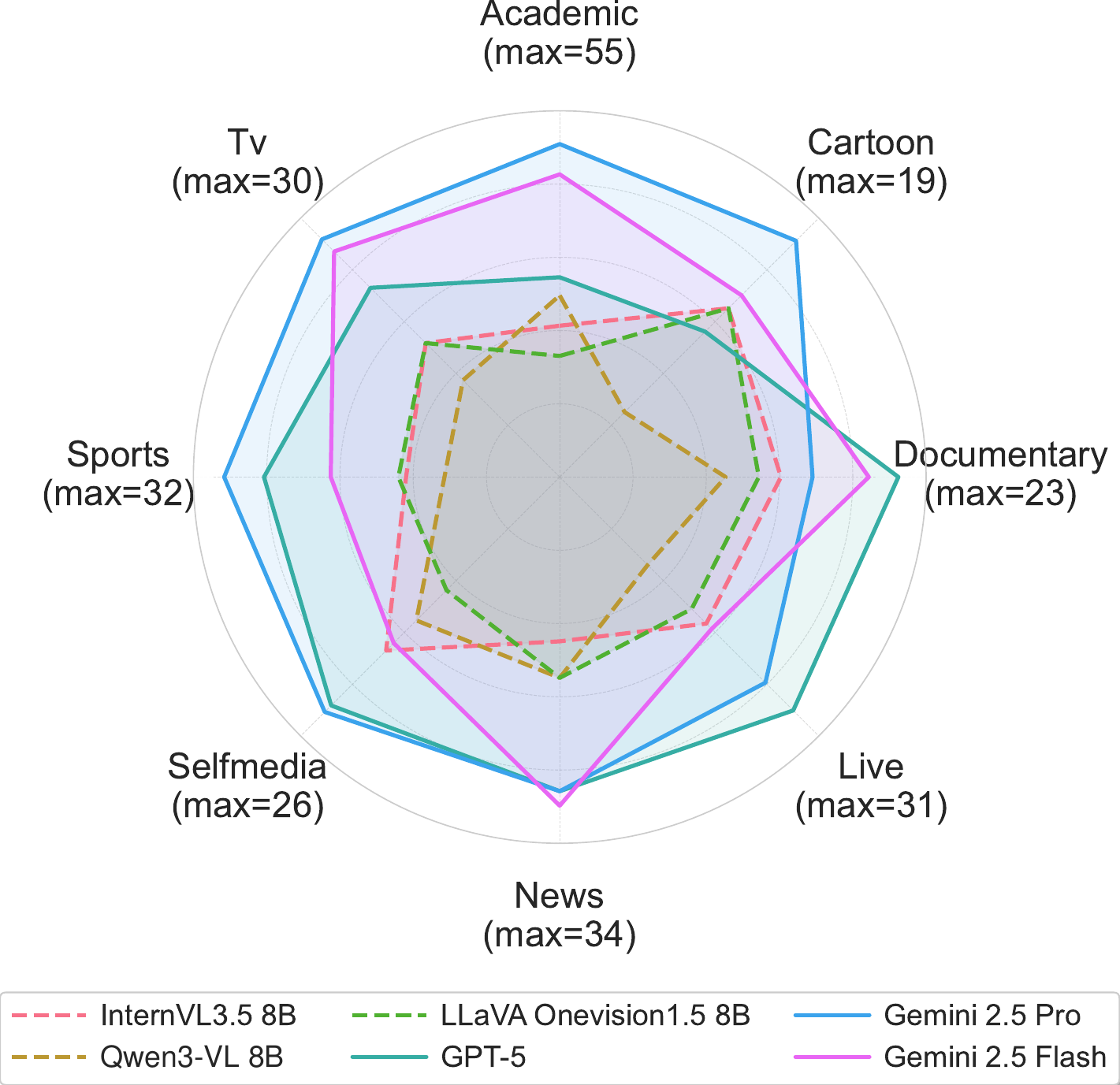}
    \caption{Counting.}
    \label{fig:radar_count_chart}
  \end{subfigure}
  \caption{Enumeration and counting accuracy across video genres.}
  \label{fig:radar_enum_count_chart}
\end{figure*}

\section{Additional Results} \label{sec:additional_results}

\noindent\textbf{Model subset used in the spotlight analyses.}
For the six-model spotlight analyses in Figure~\ref{fig:perf_by_num_clues}
(main paper), Figure~\ref{fig:radar_enum_count_chart},
Figure~\ref{fig:perf_by_query_len}, and Figure~\ref{fig:perf_by_video_len},
we use three size-matched 8B open-source models (InternVL3.5 8B,
LLaVA-OneVision1.5 8B, Qwen3-VL 8B) together with three proprietary models
(GPT-5, Gemini 2.5 Flash, Gemini 2.5 Pro), so that the open-source comparison
is at a comparable parameter count. The broader 22-model comparison in
Table~\ref{tab:enum_count_all} of the main paper additionally evaluates the
larger InternVL3.5 38B variant.

\subsection{Performance across Video Genres}

Table~\ref{tab:genre_enum_count} reports genre-level performance for a
14-model subset of the 22 models in Table~\ref{tab:enum_count_all}; the
remaining eight models are omitted from this fine-grained breakdown due
to compute constraints, while their aggregate scores in
Table~\ref{tab:enum_count_all} cover the same evaluation set.

Table~\ref{tab:genre_enum_count} compares model performance across eight video
genres: \textit{Academic}, \textit{Cartoon}, \textit{Documentary},
\textit{Live}, \textit{News}, \textit{Self-media}, \textit{Sports}, and
\textit{TV} for both the Enumeration and Counting tasks.
Across nearly all genres and tasks, proprietary models such as GPT and Gemini
consistently outperform open-source models.
However, even the strongest models achieve only around 20--30\% accuracy in
many categories, indicating that EC-Bench remains a highly challenging
benchmark for current multimodal LLMs.

\subsection{Effect of Query Length}
\begin{figure*}[!t]
\centering
\includegraphics[width=1.0\linewidth]{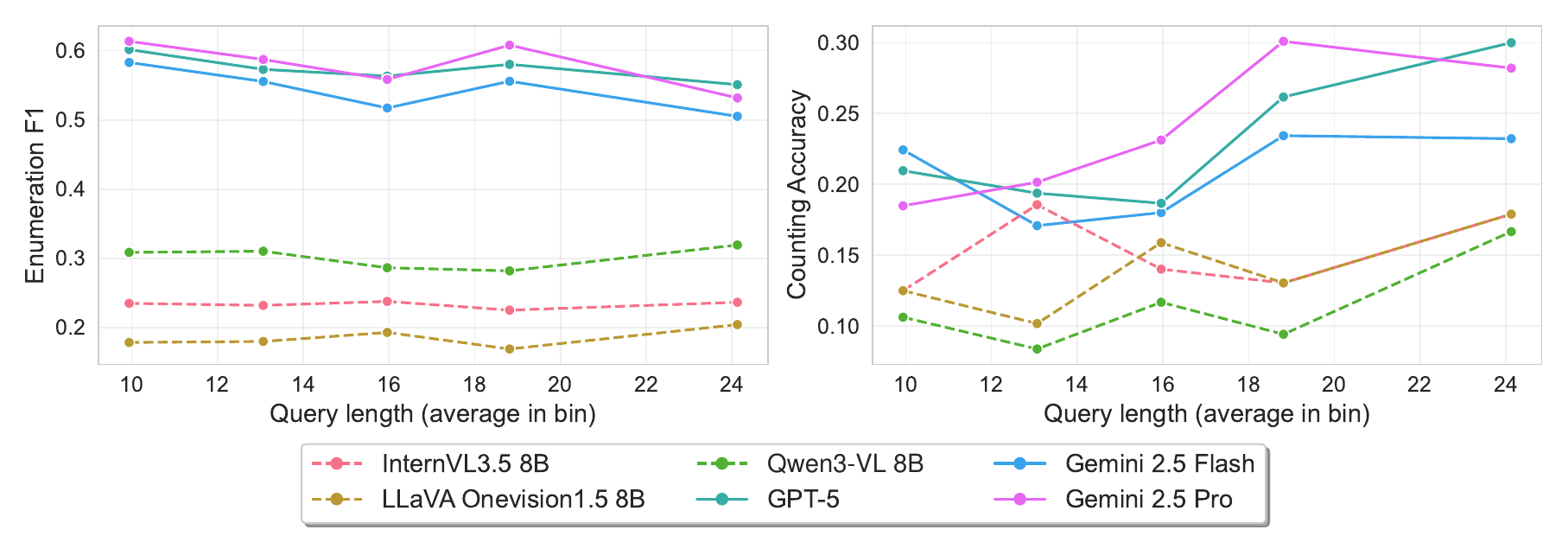}
\caption{Performance across query lengths.}
\label{fig:perf_by_query_len}
\end{figure*}

Figure~\ref{fig:perf_by_query_len} analyzes model performance across query
lengths.
Overall, longer questions do not lead to substantial performance degradation.
For Enumeration, proprietary models maintain relatively strong performance
across query lengths, while open-source models remain substantially weaker.
For Counting, proprietary models show a slight upward trend as query length
increases, whereas open-source models remain lower with only minor variations.
These results suggest that EC-Bench performance is not primarily driven by
query length, and that the performance gap between proprietary and open-source
models persists across different question lengths.

\subsection{Effect of Video Length}
\begin{figure*}[!t]
\centering
\includegraphics[width=1.0\linewidth]{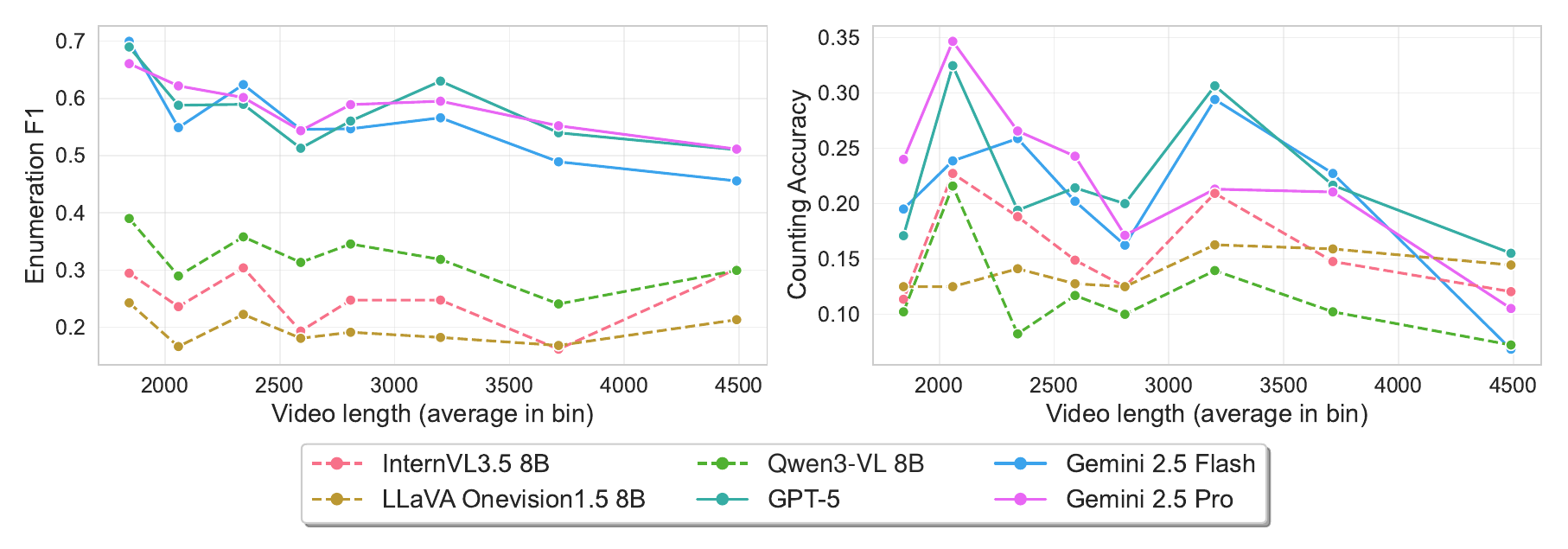}
\caption{Performance across video lengths.}
\label{fig:perf_by_video_len}
\end{figure*}

Figure~\ref{fig:perf_by_video_len} analyzes model performance across video
lengths.
Overall, performance shows a mild downward trend as videos become longer,
although results vary across bins.
For Enumeration, stronger models remain relatively stable with a gradual
decrease as video length increases, while open-source models remain
substantially lower across all lengths.
For Counting, most models exhibit slightly lower or less stable accuracy in
longer-video bins.
These results suggest that longer videos make both tasks somewhat more
challenging, while the performance gap between proprietary and open-source
models persists across video lengths.

\subsection{Failure Taxonomy} \label{app:failure_taxonomy}
To understand failures beyond aggregate accuracy, we qualitatively analyze
representative GPT-4o errors on EC-Bench and categorize them by the stage
where the error first emerges.
As summarized in Table~\ref{tab:failure_taxonomy}, incorrect counts rarely
result from isolated arithmetic mistakes; they typically arise from missed
evidence, weak temporal or modality grounding, hallucinated abstractions, or
inconsistent deduplication across long temporal contexts.
This supports our view that long-video counting should be evaluated as
evidence retrieval, grounding, and aggregation over extended videos, rather
than as standalone numerical prediction.

\begin{figure*}[t]
\centering
\begin{minipage}{0.95\textwidth}

\begin{mdframed}[
  backgroundcolor=orange!24,
  linecolor=white!20,
  linewidth=0pt,
  innerleftmargin=10pt,
  innerrightmargin=10pt,
  innertopmargin=6pt,
  innerbottommargin=6pt,
  skipabove=0pt,
  skipbelow=0pt
]
\centering
\footnotesize
\bfseries
Annotation Protocol
\end{mdframed}
\vspace{-0.7em}

\begin{mdframed}[
  backgroundcolor=orange!10,
  linecolor=white!35,
  linewidth=0.6pt,
  roundcorner=2pt,
  innerleftmargin=12pt,
  innerrightmargin=12pt,
  innertopmargin=8pt,
  innerbottommargin=8pt,
  skipabove=0pt,
  skipbelow=0pt
]

\footnotesize


\scriptsize

\begin{enumerate}[leftmargin=1.35em, itemsep=0.3em, topsep=0.2em, parsep=0pt, partopsep=0pt]

  \item \textbf{Operations.}
  Annotators update and verify the following fields:
  \begin{itemize}[leftmargin=1.5em, itemsep=0.12em, topsep=0.12em, parsep=0pt, partopsep=0pt]
      \item \textbf{Video Information}: Video URL (A), Video ID (B), Genre (C)
      \item \textbf{Question Information}: Question ID (D), Question Genre (E), Question Type (F: Counting / Enumeration), Question Text (G)
      \item \textbf{Answer and Evidence}: Pre-generated Answer (H), Pre-generated Clips (I), Revised Answer (J), Evidence Clips (K; format ``[t$_s$, t$_e$]'')
      \item \textbf{Additional Fields}: Invalid-Question Flag (L), Notes for Difficult Cases (M), Enumeration Corresponding to Counting (N)
  \end{itemize}

  \item \textbf{Step 1: Feasibility Check.}
  Annotators examine the video to determine whether the question is answerable:
  \begin{itemize}[leftmargin=1.5em, itemsep=0.12em, topsep=0.12em, parsep=0pt, partopsep=0pt]
      \item whether the question is grounded in visible content,
      \item whether the pre-generated answer matches the video,
      \item whether the assigned task type is appropriate,
      \item whether the question can be answered using video-only evidence.
  \end{itemize}
  If unanswerable, column~L is marked with (\cmark).

  \item \textbf{Step 2: Answer Verification and Revision.}
  Annotators validate the pre-generated answer (H). If it is correct, it is retained; otherwise, the corrected answer is written in column~J.

  \textit{Rules for Counting:}
  \begin{itemize}[leftmargin=1.5em, itemsep=0.12em, topsep=0.12em, parsep=0pt, partopsep=0pt]
      \item Provide only a numerical value (e.g., ``3'').
      \item Reappearances within 5 seconds count as one occurrence.
      \item For type-based questions, multiple instances of the same category count as one type.
      \item If no instance exists, write ``0''.
  \end{itemize}

  \textit{Rules for Enumeration:}
  \begin{itemize}[leftmargin=1.5em, itemsep=0.12em, topsep=0.12em, parsep=0pt, partopsep=0pt]
      \item List items using comma-separated entries.
      \item Use arrows for ordered steps (``A $\rightarrow$ B $\rightarrow$ C'').
      \item Remove modifiers and keep only the core category.
      \item If no items exist, write ``None''.
  \end{itemize}

  \textit{Ambiguous cases:}
  When unclear, annotators select the most reasonable interpretation and briefly describe the difficulty in column~M.

  \item \textbf{Step 3: Evidence Extraction.}
  Annotators list all supporting video segments in column~K:
  \begin{itemize}[leftmargin=1.5em, itemsep=0.12em, topsep=0.12em, parsep=0pt, partopsep=0pt]
      \item Second-level precision is sufficient.
      \item Include only the minimum necessary clips.
      \item For comparison questions, include evidence for all referenced entities.
      \item If one appearance suffices, extract only that segment.
      \item Merge clips separated by under 5 seconds.
  \end{itemize}

  \item \textbf{Additional Annotation Rules.}
  \begin{itemize}[leftmargin=1.5em, itemsep=0.12em, topsep=0.12em, parsep=0pt, partopsep=0pt]
      \item Use neutral placeholders when names are unknown.
      \item Remove ownership markers (e.g., ``A's car'' $\rightarrow$ ``car'').
      \item For type-based tasks, count categories rather than instances.
  \end{itemize}

\end{enumerate}

\end{mdframed}
\end{minipage}

\caption{Annotation protocol for EC-Bench.}
\label{fig:anno-protocol}
\end{figure*}

\begin{figure*}[t]
\centering
\begin{minipage}{1\linewidth}

\begin{mdframed}[
  backgroundcolor=blue!20,
  linecolor=white!20,
  linewidth=0pt,
  innerleftmargin=10pt,
  innerrightmargin=10pt,
  innertopmargin=6pt,
  innerbottommargin=6pt,
  skipabove=0pt,
  skipbelow=0pt
]
\centering
\footnotesize
\bfseries
QA Generation Prompt 
\end{mdframed}
\vspace{-0.7em}

\begin{mdframed}[
  backgroundcolor=blue!10,
  linecolor=white!35,
  linewidth=0.6pt,
  roundcorner=2pt,
  innerleftmargin=12pt,
  innerrightmargin=12pt,
  innertopmargin=8pt,
  innerbottommargin=8pt,
  skipabove=0pt,
  skipbelow=0pt
]
\renewcommand{\baselinestretch}{0.9}
\footnotesize

\scriptsize

You are a dataset annotator specializing in counting tasks for long-form videos (30 minutes or longer).
You will be given one video. Carefully examine both its visual and audio content, and create
{12 high-quality queries} following the six counting categories described below.

\vspace{0.25em}
\textbf{Six Counting Categories}

\textbf{A. Parallel Event Counting.} \emph{Definition:} Counting events that occur independently along the timeline, focusing on semantic event recognition.\\
\textit{Counting example:} ``How many times does an autonomous truck appear driving on the highway?''\\
\textit{Enumeration example:} ``List all types of flying vehicles that appear in the video.''

\vspace{0.25em}
\textbf{B. Causal Event Counting.} \emph{Definition:} Counting events that involve temporal sequence or causal relationships.\\
\textit{Counting example:} ``How many shots are exchanged in the rally beginning at 4:15?''\\
\textit{Enumeration example:} ``List the steps involved in the face-recognition payment process.''

\vspace{0.25em}
\textbf{C. Speech \& Audio Counting.} \emph{Definition:} Multi-modal counting involving speech or sound events.\\
\textit{Counting example:} ``How many times does a siren sound?''\\
\textit{Enumeration example:} ``List all industrial fields mentioned by the lecturer.''

\vspace{0.25em}
\textbf{D. Appearance Counting.} \emph{Definition:} Counting how frequently specific objects or entities visually appear.\\
\textit{Counting example:} ``How many times does a player wearing a blue uniform appear on screen?''\\
\textit{Enumeration example:} ``List all constellation-shaped creatures appearing in the video.''

\vspace{0.25em}
\textbf{E. Spatial Counting.} \emph{Definition:} Counting the number of entities present in a scene or frame.\\
\textit{Counting example:} ``How many bicycles are on the road?''\\
\textit{Enumeration example:} ``List the future medical technologies introduced in the video.''

\vspace{0.25em}
\textbf{F. Conditional Counting.} \emph{Definition:} Advanced reasoning requiring both spatial and temporal conditions.\\
\textit{Counting example:} ``How many shots occur inside the penalty area during the second half?''\\
\textit{Enumeration example:} ``List all points of change compared with 50 years ago.''

\vspace{0.25em}
\textbf{Answer Format Rules}

\textbf{1. Counting questions.} \emph{Definition:} Numerical answers such as counts or quantities. \emph{Rules:}
\begin{itemize}[leftmargin=1.4em,itemsep=0.12em,topsep=0.12em,parsep=0pt,partopsep=0pt]
  \item Answer must be a number only (units optional).
  \item Keep expressions concise---no full sentences.
  \item Use commas if multiple values are needed.
\end{itemize}

\textbf{2. Enumeration questions.} \emph{Definition:} Listing multiple items or steps. \emph{Rules:}
\begin{itemize}[leftmargin=1.4em,itemsep=0.12em,topsep=0.12em,parsep=0pt,partopsep=0pt]
  \item Items separated by commas.
  \item Use ``$\rightarrow$'' for ordered sequences.
  \item Do not add explanations.
\end{itemize}

\vspace{0.25em}
\textbf{Mandatory Global Requirements}
\begin{itemize}[leftmargin=1.4em,itemsep=0.12em,topsep=0.12em,parsep=0pt,partopsep=0pt]
  \item Must require understanding of the entire video.
  \item Must require visual and/or audio evidence.
  \item Prefer questions that require multiple evidence clips.
  \item Must remain fully objective.
  \item Except for Category C, rely on visual information only.
  \item Cannot depend on external world knowledge.
  \item Must be concise, single-line questions.
\end{itemize}

\vspace{0.25em}
\textbf{Template Questions for Sports Videos}\\
\textit{\textless QA Examples\textgreater}

\vspace{0.25em}
\textbf{Output Format (JSON Only)}
\vspace{-0.85em}
\begin{verbatim}
{
  "queries": [
    {"category": "A", "tag": "Counting",
     "query": "...", "answer": "..."},
    {"category": "A", "tag": "Enumeration",
     "query": "...", "answer": "..."},
    ...
  ]
}
\end{verbatim}

\end{mdframed}
\end{minipage}
\caption{QA generation prompt for long-video Enumeration and Counting tasks.}
\label{fig:qa_generation_prompt}
\end{figure*}

\begin{figure*}[t]
\centering

\begin{mdframed}[
  backgroundcolor=green!24,
  linecolor=white!20,
  linewidth=0pt,
  innerleftmargin=10pt,
  innerrightmargin=10pt,
  innertopmargin=6pt,
  innerbottommargin=6pt,
  skipabove=0pt,
  skipbelow=0pt
]
\centering
\footnotesize
\bfseries
Enumeration and Counting QA Prompt
\end{mdframed}
\vspace{-0.8em}

\begin{minipage}{1\textwidth}
\begin{mdframed}[
  backgroundcolor=green!6,
  linecolor=white!35,
  linewidth=0.5pt,
  roundcorner=2pt,
  innerleftmargin=10pt,
  innerrightmargin=10pt,
  innertopmargin=6pt,
  innerbottommargin=6pt,
  skipabove=0pt,
  skipbelow=0pt
]

\begingroup
\renewcommand{\baselinestretch}{0.90}
\footnotesize
\selectfont


\scriptsize

You are an expert annotator specializing in counting tasks for long-form videos (30+ minutes). Please analyze the given sampled video frames and audio transcription in detail, generate accurate answers for all the following queries, and record the evidence clips that support each answer.

\vspace{0.25em}
\textbf{\{queries\_section\}}

\textbf{\{frames\_section\}}

\textbf{\{transcription\_section\}}

\vspace{0.2em}
\textbf{Answer Rules}

\textbf{1. Counting Type}\\
\emph{Definition:} Questions that require numerical answers such as frequency, number of people, or percentage of similar entities or events.\\
\emph{Recording Rules:}
\begin{itemize}[leftmargin=1.25em,itemsep=0.05em,topsep=0.05em,parsep=0pt,partopsep=0pt]
  \item Numbers only
  \item No sentences---record only numbers
  \item For multiple values: comma-separated (e.g., ``2, 5, 7'')
  \item \textbf{5-second rule}: If the same subject reappears within 5 seconds, count it as a single occurrence
\end{itemize}

\vspace{0.25em}
\textbf{2. Enumeration Type}\\
\emph{Definition:} Questions that require listing multiple entities or procedures.\\
\emph{Recording Rules:}
\begin{itemize}[leftmargin=1.25em,itemsep=0.05em,topsep=0.05em,parsep=0pt,partopsep=0pt]
  \item List items separated by commas
  \item For ordered (causal/procedural) sequences: connect using ``$\rightarrow$''
  \item No explanations---items only
  \item Omit ownership names (``A's car'' $\rightarrow$ ``car'')
\end{itemize}

\vspace{0.25em}
\emph{Recording Examples:}
\begin{itemize}[leftmargin=1.25em,itemsep=0.05em,topsep=0.05em,parsep=0pt,partopsep=0pt]
  \item ``bat, glove, ball''
  \item ``Player A, Player B, Player C''
  \item ``forging $\rightarrow$ polishing $\rightarrow$ decoration $\rightarrow$ completion''
  \item ``facial recognition $\rightarrow$ amount confirmation $\rightarrow$ approval $\rightarrow$ payment completion''
\end{itemize}

\vspace{0.25em}
\textbf{Analysis Instructions}
\begin{enumerate}[leftmargin=1.25em,itemsep=0.08em,topsep=0.08em,parsep=0pt,partopsep=0pt]
  \item Carefully examine all provided video frames in chronological order
  \item Use the audio transcription to understand context and dialog
  \item Combine visual and audio information to generate accurate answers
  \item Pay attention to temporal relationships between frames when counting or enumerating
  \item Identify precise video clips that support each answer
\end{enumerate}

\vspace{0.25em}
\textbf{Output Format}

Please provide answers in the following JSON format:

{\ttfamily\scriptsize
\noindent\{\\
\hspace*{1em}"results": [\\
\hspace*{2em}\{\\
\hspace*{3em}"query\_id": 1,\\
\hspace*{3em}"answer": "accurate answer",\\
\hspace*{3em}"clips": [\\
\hspace*{4em}["00:04:12", "00:07:23"],\\
\hspace*{4em}["00:12:12", "00:12:56"]\\
\hspace*{3em}]\\
\hspace*{2em}\}\\
\hspace*{1em}]\\
\}
}

\vspace{0.25em}
\textbf{Important Notes:}
\begin{itemize}[leftmargin=1.25em,itemsep=0.05em,topsep=0.05em,parsep=0pt,partopsep=0pt]
  \item Choose the correct answer format:
    \begin{itemize}[leftmargin=1.2em,itemsep=0.03em,topsep=0.03em,parsep=0pt,partopsep=0pt]
      \item Counting $\rightarrow$ numbers only
      \item Enumeration $\rightarrow$ comma-separated items or ``$\rightarrow$''
    \end{itemize}
  \item \texttt{clips} must list all evidence intervals as [start, end] timestamp pairs.
  \item Answers must adhere strictly to all rules above.
  \item Use both visual frame information and audio transcription.
\end{itemize}

\endgroup
\end{mdframed}
\end{minipage}
\caption{Enumeration and counting QA prompt for long videos.}
\label{fig:qa_prompt}
\end{figure*}

\begin{figure*}[t]
\centering
\begin{minipage}{1\textwidth}

\begin{mdframed}[
  backgroundcolor=green!24,
  linecolor=white!20,
  linewidth=0pt,
  innerleftmargin=10pt,
  innerrightmargin=10pt,
  innertopmargin=6pt,
  innerbottommargin=6pt,
  skipabove=0pt,
  skipbelow=0pt
]
\centering
\footnotesize
\bfseries
Enumeration-First Counting Prompt
\end{mdframed}
\vspace{-0.7em}

\begin{mdframed}[
  backgroundcolor=green!6,
  linecolor=gray!35,
  linewidth=0pt,
  roundcorner=2pt,
  innerleftmargin=10pt,
  innerrightmargin=10pt,
  innertopmargin=6pt,
  innerbottommargin=6pt,
  skipabove=0pt,
  skipbelow=0pt
]

\begingroup
\renewcommand{\baselinestretch}{0.90}
\footnotesize
\selectfont



You are an expert annotator specializing in counting tasks for long-form videos (30+ minutes).
Follow an \emph{enumerate-first, then count} workflow and record the evidence clips that support your answer.

\vspace{0.25em}
\textbf{\{query\_section\}}

\textbf{\{frames\_section\}}

\textbf{\{transcription\_section\}}

\vspace{0.2em}
\textbf{Counting Answer Rules}
\begin{itemize}[leftmargin=1.25em,itemsep=0.05em,topsep=0.05em,parsep=0pt,partopsep=0pt]
  \item Numbers only (e.g., ``2'' or ``2, 5'' for multiple values)
  \item Keep answers concise; do not write sentences
  \item Separate multiple values using commas
  \item \textbf{5-second rule}: repeated appearances of the same subject within 5 seconds count as one
  \item Use both visual frames and transcript cues to confirm instances
\end{itemize}

\vspace{0.25em}
\textbf{Step-by-step Instructions}
\begin{enumerate}[leftmargin=1.25em,itemsep=0.08em,topsep=0.08em,parsep=0pt,partopsep=0pt]
  \item Enumerate every distinct instance relevant to the query
  \item Describe each instance with distinguishing details (appearance, action, timestamp hints)
  \item Deduplicate if necessary and compute the final count
  \item Link each instance to precise evidence clips
  \item Do not guess—return only answers supported by provided data
\end{enumerate}

\vspace{0.25em}
\textbf{Output Format}

Return the answer using the following JSON structure:

{\ttfamily\scriptsize
\noindent\{\\
\hspace*{1em}"enumeration": [\\
\hspace*{2em}"Instance description 1",\\
\hspace*{2em}"Instance description 2"\\
\hspace*{1em}],\\
\hspace*{1em}"answer": "2",\\
\hspace*{1em}"clips": [\\
\hspace*{2em}["00:05:10", "00:05:35"],\\
\hspace*{2em}["00:12:42", "00:13:05"]\\
\hspace*{1em}]\\
\}
}

\vspace{0.25em}
\textbf{Notes}
\begin{itemize}[leftmargin=1.25em,itemsep=0.05em,topsep=0.05em,parsep=0pt,partopsep=0pt]
  \item \texttt{enumeration}: list of all counted instances
  \item \texttt{answer}: final numeric count (string, no units)
  \item \texttt{clips}: supporting evidence segments
\end{itemize}

\endgroup
\end{mdframed}
\end{minipage}

\caption{Enumeration-first prompt for long-form video counting.}
\label{fig:enum_first_prompt}

\end{figure*}

\begin{figure*}[t]
\centering
\begin{minipage}{1\textwidth}

\begin{mdframed}[
  backgroundcolor=red!16,
  linecolor=white!20,
  linewidth=0pt,
  innerleftmargin=10pt,
  innerrightmargin=10pt,
  innertopmargin=6pt,
  innerbottommargin=6pt,
  skipabove=0pt,
  skipbelow=0pt
]
\centering
\footnotesize
\bfseries
LLM-as-Judge Prompt
\end{mdframed}
\vspace{-0.7em}

\begin{mdframed}[
  backgroundcolor=red!4,
  linecolor=white!35,
  linewidth=0.5pt,
  roundcorner=2pt,
  innerleftmargin=10pt,
  innerrightmargin=10pt,
  innertopmargin=6pt,
  innerbottommargin=6pt,
  skipabove=0pt,
  skipbelow=0pt
]

\begingroup
\renewcommand{\baselinestretch}{0.90}
\footnotesize
\selectfont


You are an evaluation model designed to compare two enumeration-style answers.
Your goal is to judge correctness, completeness, and semantic equivalence between
a ground-truth enumeration and a generated answer.

\vspace{0.25em}
\textbf{Evaluation Tasks}
\begin{itemize}[leftmargin=1.25em,itemsep=0.06em,topsep=0.06em,parsep=0pt,partopsep=0pt]
  \item Identify the essential items/entities present in the ground truth
  \item Determine which items appear in both answers (true positives)
  \item Identify incorrectly added items in the generated answer (false positives)
  \item Identify missing ground-truth items (false negatives)
  \item Treat semantically equivalent expressions as the same entity
\end{itemize}

\vspace{0.25em}
\textbf{Handling Synonyms and Paraphrases}
\begin{itemize}[leftmargin=1.25em,itemsep=0.05em,topsep=0.05em,parsep=0pt,partopsep=0pt]
  \item Merge expressions referring to the same real-world object or event
  \item Combine alternate names, abbreviations, and titles  
        (e.g., ``fire truck'' = ``fire engine''; ``goalkeeper'' = ``goalie'')
  \item Ignore modifiers unless they fundamentally change the identity
\end{itemize}

\vspace{0.25em}
\textbf{Required Output Format}

Produce evaluation results using the following JSON structure:

{\ttfamily\scriptsize
\noindent\{\\
\hspace*{1em}"tp\_items": ["string", ...],\\
\hspace*{1em}"fp\_items": ["string", ...],\\
\hspace*{1em}"fn\_items": ["string", ...],\\
\hspace*{1em}"reasoning": "brief explanation",\\
\hspace*{1em}"confidence": 0.0\\
\}
}

\vspace{0.25em}
\textbf{Rules}
\begin{itemize}[leftmargin=1.25em,itemsep=0.05em,topsep=0.05em,parsep=0pt,partopsep=0pt]
  \item Arrays must always be present (empty allowed)
  \item ``confidence'' is a float between 0 and 1
  \item Output JSON only—no commentary or markdown fencing
\end{itemize}

\vspace{0.25em}
\textbf{Inputs}
\begin{itemize}[leftmargin=1.25em,itemsep=0.05em,topsep=0.05em,parsep=0pt,partopsep=0pt]
  \item True answer: \texttt{\{true\_answer\}}
  \item Generated answer: \texttt{\{generated\_answer\}}
\end{itemize}

\endgroup
\end{mdframed}
\end{minipage}

\caption{Prompt for evaluating enumeration-style answers using an LLM-as-Judge.}
\label{fig:llm_judge_prompt}

\end{figure*}

\end{document}